\documentclass[10pt,journal,compsoc]{IEEEtran}
\ifCLASSOPTIONcompsoc
\else
\fi

\ifCLASSINFOpdf
\else
\fi
\usepackage{hyperref}
\usepackage{forest}
\usepackage[noadjust]{cite}
\usepackage{multirow}
\usepackage{amsmath}
\usepackage{pifont}% http://ctan.org/pkg/pifont
\usepackage{caption}% http://ctan.org/pkg/caption
\usepackage{enumitem} 
\begin{document}

\title{AMMU : A Survey of Transformer-based Biomedical Pretrained Language Models}

\author{Katikapalli~Subramanyam~Kalyan,
        Ajit~Rajasekharan,
        and~Sivanesan~Sangeetha% <-this % stops a space
\IEEEcompsocitemizethanks{\IEEEcompsocthanksitem K.S.Kalyan is with the Department
of Computer Applications, National Institute of Technology Trichy, Trichy,
Tamil Nadu, India, 620015.\protect\\
E-mail: kalyan.ks@yahoo.com, Website: \url{https://mr-nlp.github.io}
\IEEEcompsocthanksitem Ajit Rajasekharan is with the Nference.ai as CTO, Cambridge, MA, USA, 02142.
\IEEEcompsocthanksitem S.Sangeetha  is with the Department of Computer Applications, National Institute of Technology Trichy, Trichy, Tamil Nadu, India, 620015..}% <-this % stops an unwanted space
\thanks{Preprint under review - The paper is named (AMMU) in the memory of one of the close friends of K.S.Kalyan (\url{https://mr-nlp.github.io}).}}

\IEEEtitleabstractindextext{%
\begin{abstract}

Transformer-based pretrained language models (PLMs) have started a new era in modern natural language processing (NLP). These models combine the power of transformers, transfer learning, and self-supervised learning (SSL). Following the success of these models in the general domain, the biomedical research community has developed various in-domain PLMs starting from BioBERT to the latest BioELECTRA and BioALBERT models. We strongly believe there is a need for a survey paper that can provide a comprehensive survey of various transformer-based biomedical pretrained language models (BPLMs).  In this survey, we start with a brief overview of foundational concepts like self-supervised learning, embedding layer and transformer encoder layers. We discuss core concepts of transformer-based PLMs like pretraining methods, pretraining tasks, fine-tuning methods, and various embedding types specific to biomedical domain. We introduce a taxonomy for transformer-based BPLMs and then discuss all the models. We discuss various challenges and present possible solutions. We conclude by highlighting some of the open issues which will drive the research community to further improve transformer-based BPLMs. The list of all the publicly available transformer-based BPLMs along with their links is provided at \underline{\textbf{\url{https://mr-nlp.github.io/posts/2021/05/transformer-based-biomedical-pretrained-language-models-list/}}}.
\end{abstract}

}

\maketitle

\IEEEdisplaynontitleabstractindextext

% \fi
\IEEEpeerreviewmaketitle

\tableofcontents
\section{Introduction}
\begin{figure*}[h]
\begin{center}
\includegraphics[width=14cm, height=12cm]{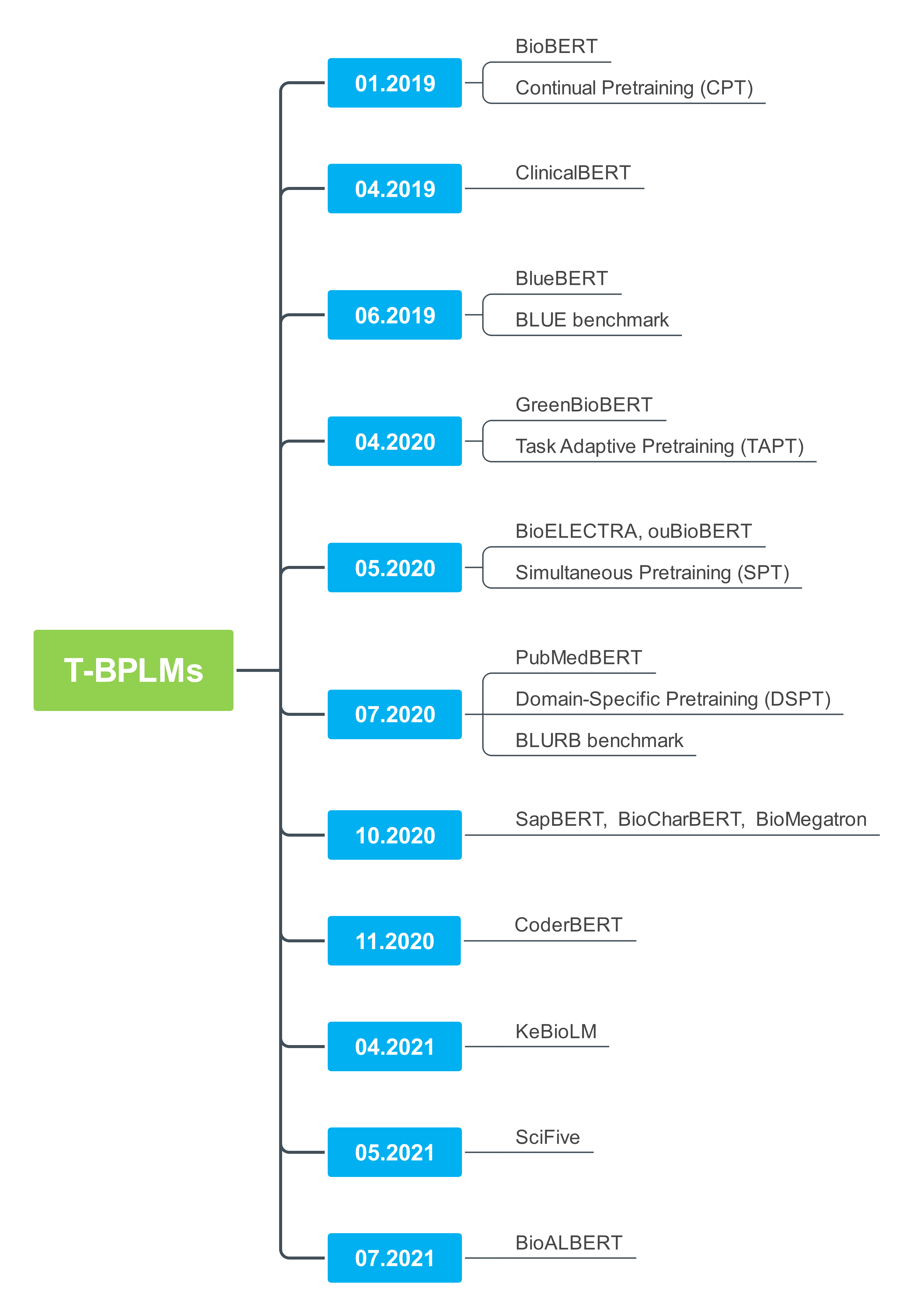}
\caption{\label{ammu-timeline} Key milestones in T-BPLMs } 
\end{center}
\end{figure*}
\IEEEPARstart{T}{ransformer}\cite{vaswani2017attention} based PLMs like BERT \cite{devlin2019bert}, RoBERTa \cite{liu2019roberta}, T5 \cite{raffel2020exploring} have started a new era in modern NLP. These models combine the power of transformers, transfer learning, and self-supervised learning. Transformers use self-attention which can be run in parallel and can model long-range relationships with ease. In transfer learning \cite{pan2009survey}, knowledge gained by the model in the source task is transferred to the target task. For example, computer vision models are trained over large labeled datasets, and then these pretrained models are used in similar tasks where the labeled datasets are small \cite{krizhevsky2012imagenet,szegedy2015going}. The main advantages of pretrained models are a) they learn language representations that are useful across tasks and b) no need to train the downstream models from scratch. However, in NLP, it is quite expensive and difficult to obtain such large, annotated datasets. So, transformer-based PLMs are pretrained over large unlabeled text data using self-supervised learning. Self-supervised learning is in between supervised and unsupervised learning. Supervised learning requires human-annotated instances while unsupervised learning does not require any labeled instances. Self-supervised learning relies on labels like supervised and semi-supervised learning. However, these labels are not human assigned but created automatically by using the relationships between various sections of the input data. Once the model is pre-trained over large volumes of text, it can be used in various downstream tasks by fine-tuning after adding task-specific layers \cite{devlin2019bert}. 

\begin{figure*}[h]
\begin{center}
\includegraphics[width=18cm, height=10cm]{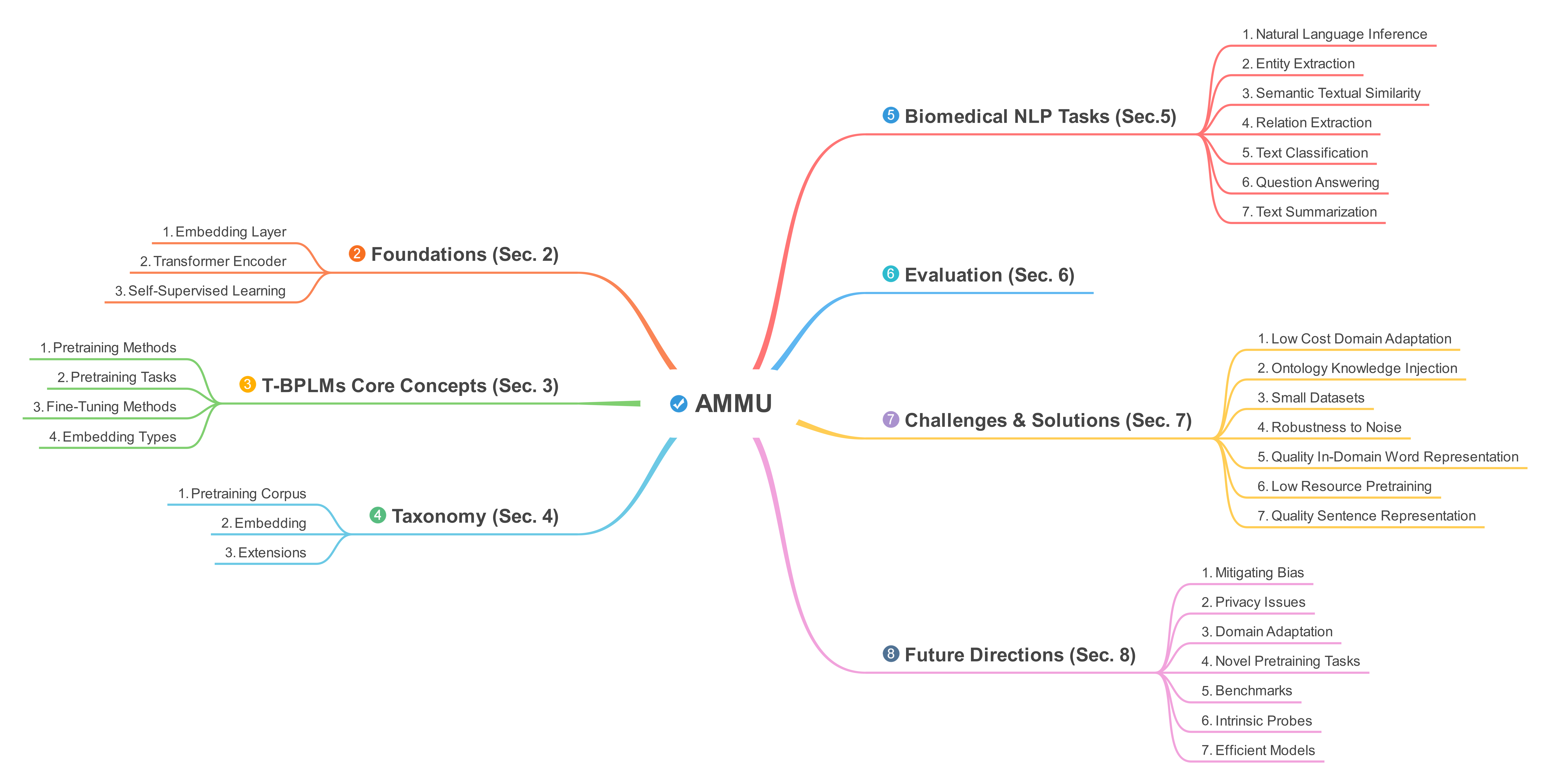}
\caption{\label{ammu-summary} Summmary of AMMU survey paper } 
\end{center}
\end{figure*}

In the initial days, NLP systems are mostly rule-based. The development of rule-based systems is quite difficult as it requires significant human intervention in the form of domain expertise to frame the rules. It is required to reframe the rules with even with a small change in the input data which makes it expensive and laborious. Machine learning systems to some extent brought flexibility in developing NLP systems. Machine learning systems learn the rules during training and thereby avoids the laborious process of manual rule framing. However, the main drawback in machine learning models is the requirement of feature engineering which again requires domain expertise. With the development of various deep learning models like convolutional neural networks (CNNs) and recurrent neural networks (RNNs) which can learn features automatically and better hardware like GPUs, NLP researchers shifted to deep learning models with dense word vectors as input \cite{blunsom2014convolutional,liu2016recurrent}. Traditional text representation methods like tf-idf and one-hot vectors are high-dimensional which demand more computational resources. Moreover, these representations are unable to encode syntactic and sematic information. This requirement of low-dimensional text vectors which can also encode language information leads to the development of embedding models like Word2Vec \cite{mikolov2013efficient}, Glove \cite{pennington2014glove}. As these models cannot encode sub-word information and suffer from the out of vocabulary (OOV) problem, FastText \cite{bojanowski2017enriching} is proposed. Some of the drawbacks of using CNN or RNN with dense word vectors as input are a) Embeddings models like Word2Vec, Glove, and FastText are based on shallow neural networks. Shallow neural networks with only two or three layers are unable to capture more language information into word vectors. Being context insensitive further limits the quality of these word vectors. b) Even though word embeddings are pre-trained on text corpus, the parameters of models like CNN and RNN are randomly initialized and learned during model training. Learning model parameters from scratch requires a large number of training instances.

Self-attention computes the representation of every token in the input based on its interaction with every token in the input. As a result, the self-attention mechanism can better handle long distance word relationships compared to CNN and RNN \cite{vaswani2017attention,qiu2020pre,kalyan2021ammus}. Moreover, transformers can learn complex language information by applying self-attention layers iteratively i.e., by using a stack of self-attention layers. Transformers with self-attention as the core component have become the primary choice of architecture for pretrained language models in NLP. Transformer-based PLMs like BERT \cite{devlin2019bert}, RoBERTa \cite{liu2019roberta}, ALBERT \cite{lan2019albert}, T5 \cite{raffel2020exploring} achieved tremendous success in many of the NLP tasks. These models eliminate the requirement of training a downstream model from scratch. With the success of these models, pretraining the model on large volumes of text and then fine-tuning it  on task-specific datasets has become a standard approach in modern NLP. Following the success of transformer-based PLMs in the general domain, biomedical NLP researchers have developed models like BioBERT \cite{lee2020biobert}, ClinicalBERT \cite{alsentzer2019publicly}, and BlueBERT \cite{peng2019transfer}. All these models are obtained by further pretraining general BERT on biomedical texts except ClinicalBERT which is initialized from BioBERT. 

Lee et al. \cite{lee2020biobert} proposed BioBERT in January 2019 and it is the first transformer-based BPLM. After that, number of models are proposed like ClinicalBERT \cite{alsentzer2019publicly}, ClinicalXLNet \cite{huang2020clinical}, BlueBERT \cite{peng2019transfer}, PubMedBERT \cite{gu2020domain}, ouBioBERT \cite{wada2020pre}. Since BioBERT, around 40+ BPLMs are proposed to push the state-of-the-art in various biomedical NLP tasks. Figure \ref{ammu-timeline} summarizes key milestones in transformer-based BPLMs. Transformer-based BPLMs have become the first choice for any task in biomedical NLP. However, there is no survey paper that presents the recent trends in the transformer-based PLMs in biomedical NLP.

Currently, there are three survey papers that provide a comprehensive review of embeddings in the biomedical domain and three survey papers that provide a comprehensive review of transformer-based PLMs in the general domain. The survey paper written by Kalyan and Sangeetha \cite{kalyan2020secnlp} is the first comprehensive survey on embeddings in biomedical NLP. This paper a) classify and compare various biomedical corpora b) present a brief overview of various context insensitive embedding models and compare them c) classify and explain various biomedical embeddings d) present solutions to various challenges in biomedical embeddings. The survey papers written by Chiu and Baker \cite{chiu2020word}, Khattak et al. \cite{khattak2019survey} also present the same contents differently. All these three survey papers provide information mostly on context insensitive biomedical embeddings with very little emphasis on transformer-based BPLMs. The paper by Wang et al. \cite{wang2018comparison} provides empirical evaluation of word embeddings trained from various corpora.  The survey papers written by Qiu et al. \cite{qiu2020pre}, Liu et al. \cite{liu2020survey} and Kalyan et al. \cite{kalyan2021ammus} present a review of various transformer-based PLMs in the general domain only. So, we strongly believe there is a need for a survey paper that presents the recent trends related to transformer-based BPLMs (T-BPLMs). Figure \ref{ammu-summary} summarizes the contents of this survey paper.

\subsection{Literature Search and Selection}

\begin{figure*}[h]
\begin{center}
\includegraphics[width=14cm, height=8cm]{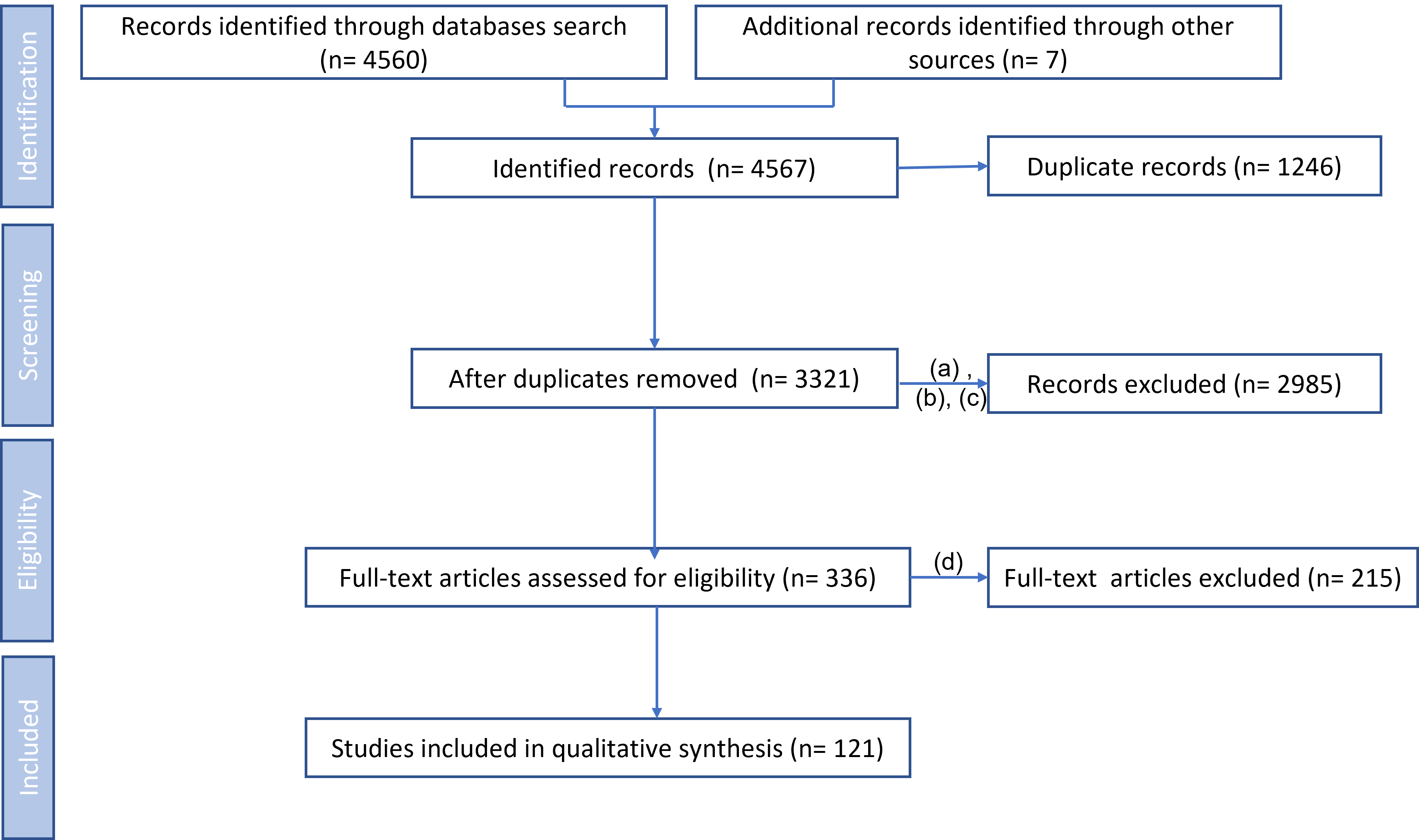}
\caption{\label{ammu-prism} PRISM flowchart for literature selection } 
\end{center}
\end{figure*}

Figure \ref{ammu-prism} shows the PRISMA  flow chart for literature search and selection. For the literature survey, we searched in databases like PubMed, ACM Digital Library, IEEE Xplore, ACL Web Anthology, Google Scholar and ScienceDirect. The first transformer-based BPLM i.e., BioBERT was released in January 2019. So, we gathered articles published in between January 2019 and July 2021. For the literature search, we initially used keywords like “biomedical pretrained models”, “clinical pretrained models”, “BioBERT”, “PubMedBERT”, “BlueBERT”, “ClinicalBERT”, “transformer-based language models” and “in-domain pretrained models”. We iteratively added new keywords from the gathered articles and finally arrived at this list of keywords "biomedical pretrained models", "clinical pretrained models", "BioBERT", "PubMedBERT", "BlueBERT", "ClinicalBERT", "transformer-based language models", "in-domain pretrained models", “BioELECTRA”, “BioALBERT”, “BLUE benchmark”, “BLURB benchmark”, “transformers”, “domain-specific pretrained models”, “medical language models”, “multi-modal pretrained models”.  Finally, we collected around 4567 articles out of which 1246 articles were duplicate. After excluding the duplicate and irrelevant articles, there were 121 articles.  We considered an article as irrelevant based on the following
\begin{enumerate} [label=(\alph*)]
    \item  article is not related to natural language processing (321 articles)
    \item article is related to natural language processing but not related to biomedical domain (2509 articles)
    \item article is related to biomedical domain but the approach is mainly based on context insensitive embeddings models and cited T-BPLMs papers in future work (155 articles).
    \item  article is related to biomedical domain and approach is based on T-BPLMs but the approach involves mere application of T-BPLMs without much novelty (215 articles).

\end{enumerate}

The highlights of this survey paper are
\begin{itemize}
    \item First survey paper to present the recent trends in transformer-based BPLMs.
    \item We present a brief overview of various foundational concepts like embedding layer, transformer encoder layer and self-supervised learning (Section \ref{foundations-sec}).
    \item We explain various core concepts related to transformer-based BPLMs like pretraining methods, pretraining tasks, fine-tuning methods, and embeddings. We discuss each concept in detail, classify and compare various methods in each (Section \ref{core-concepts-sec}). 
    \item We present a taxonomy of transformer-based BPLMs and present a brief overview of all the models (Section \ref{taxonomy-sec}).
   \item We explain how transformer-based BPLMs are applied in various biomedical NLP tasks (Section \ref{nlptasks-sec}).
    \item We present solutions to some of the challenges like low-cost domain adaptation, small biomedical datasets, ontology knowledge injection, robustness to noise, quality in-domain word representations, quality sequence representation and pretraining using less in-domain corpora (Section \ref{challenges-sec}).
    \item  We discuss possible future directions which will drive the researchers to further enhance transformer-based BPLMs (Section \ref{future-sec}).
\end{itemize}

\section{Foundations }
\label{foundations-sec}

\begin{figure}[h]
\begin{center}
\includegraphics[width=5cm, height=7cm]{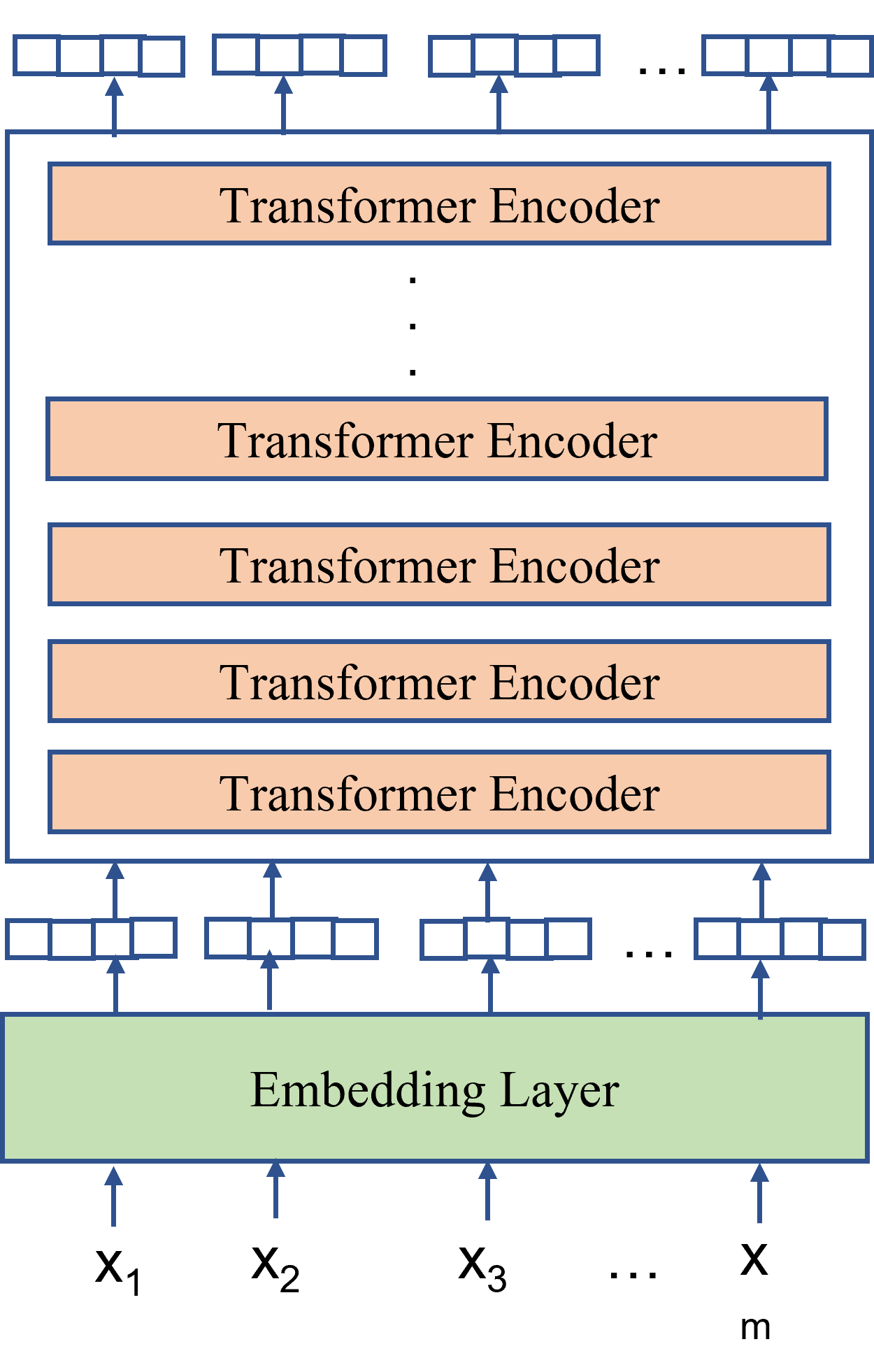}
\caption{\label{ammu-transformer} T-PLM like BERT and RoBERTa } 
\end{center}
\end{figure}

In general, the core components of transformer-based PLMs like BERT and RoBERTa are embedding and transformer encoder layers (refer Figure \ref{ammu-transformer}). The embedding layer takes input tokens and returns a vector for each. The embedding layer has three or more sub-layers each of which provides a vector of specific embedding type for each of the input tokens. The final input vector for each token is obtained by summing all the vectors of each embedding type. The transformer encoder layer enhances each input token vector by encoding global contextual information using the self-attention mechanism. By applying a sequence of such transformer encoder layers, the model can encode complex language information in the input token vectors. 

\subsection{Embedding Layer}

\begin{figure}[h]
\begin{center}
\includegraphics[width=8cm, height=4cm]{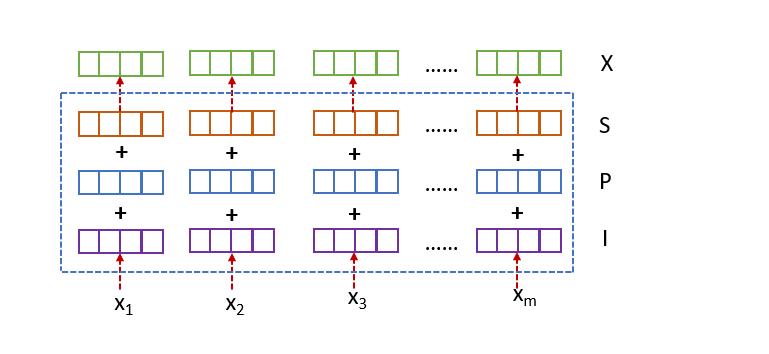}
\caption{\label{ammu-embedding-matrix} Final input vectors obtained by summing all the three vectors.} 
\end{center}
\end{figure}

Usually, the embedding layer consists of three sub-layers with each sub-layer representing a particular embedding type. In some models, there are more than three also. For example, embedding layer of BERT-EHR \cite{li2020behrt} contains code, position, segment, age and gender embeddings. A detailed description of various embedding types is presented in Section \ref{embeddings-ssec}. The first sub-layer converts input tokens to a sequence of vectors while the other two sub-layers provide auxiliary information like position and segmentation. The first sub-layer can be char, sub-word, or code embedding based. For example, BioCharBERT \cite{el2020characterbert} uses CharCNN \cite{kim2016character} on the top of character embeddings, BERT uses WordPiece \cite{wu2016google} embeddings while BEHRT \cite{li2020behrt}, MedBERT \cite{rasmy2021med} and BERT-EHR \cite{meng2021bidirectional} models use code embeddings. Unlike BioCharBERT and BERT models, the input for BEHRT, MedBERT, BERT-EHR models is patient visits where each patient visit is expressed as a sequence of codes. The final input representation $X$ for the given input tokens $\{x_1, x_2, … x_n\}$is obtained by adding the embeddings from the three sub-layers (for simplicity, we have included only three embedding types – refer Figure \ref{ammu-embedding-matrix}).

\begin{equation}
    X=I+P+S
\end{equation}
Where $X \in R^{n \: \times \: e}$  represents final input embeddings matrix and $I \in R^{n \: \times \: e}$, $P \in R^{n \: \times \: e}$  and $S \in R^{n \: \times \: e}$ represents the three embedding type matrices. Here $n$ represents length of input sequence and $e$ represents embedding size.

\subsection{Transformer Encoder}

\begin{figure}[h]
\begin{center}
\includegraphics[width=5cm, height=7cm]{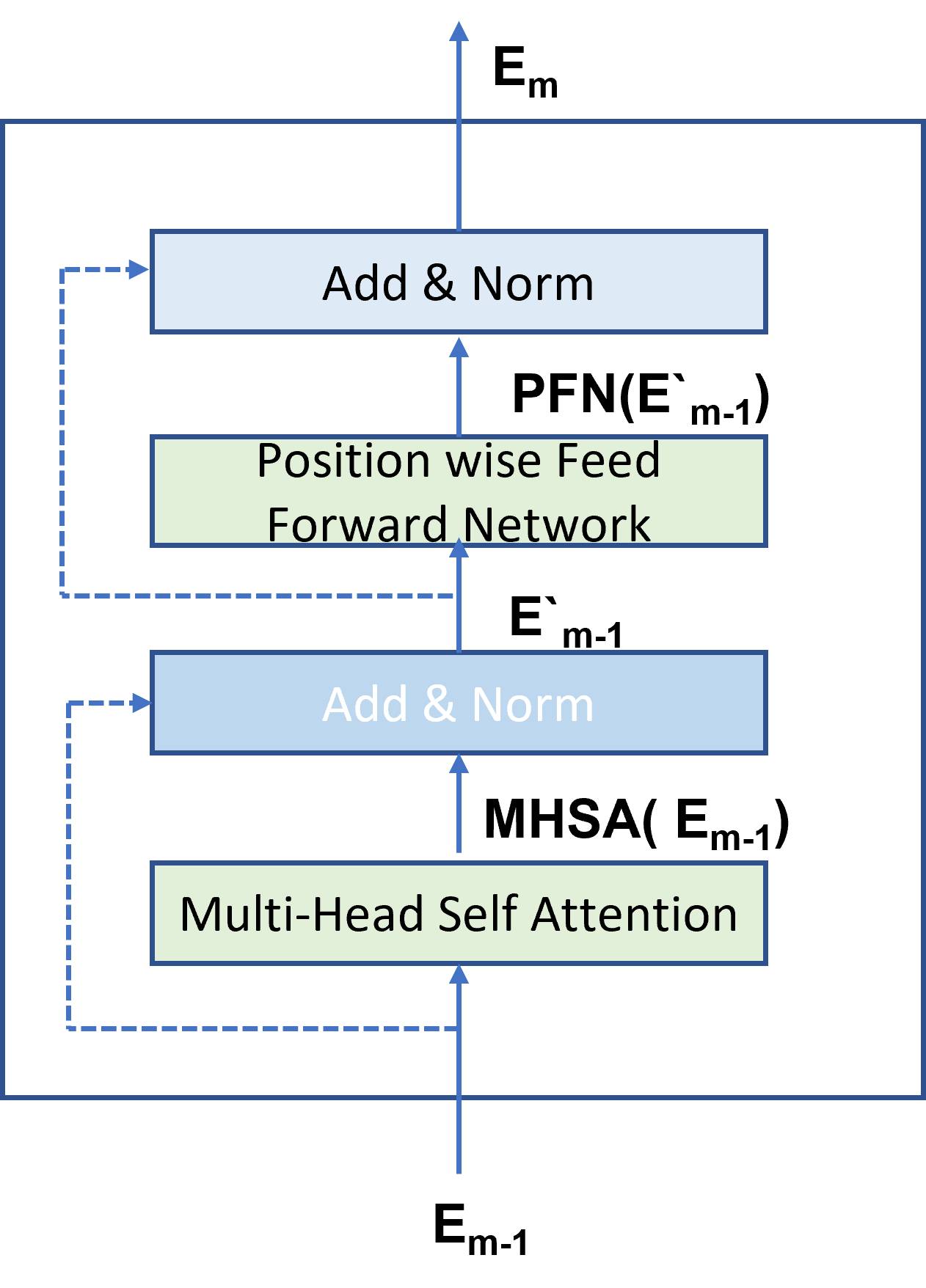}
\caption{\label{ammu-encoder} Transformer Encoder } 
\end{center}
\end{figure}

Multi-Head Self Attention (MHSA), Position-wise Feed Forward Network (PFN), Add and Norm constitutes a transformer encoder layer (refer Figure \ref{ammu-encoder}). MHSA applies self-attention (SA) multiple times independently to relate each token to all the tokens in the input sequence, while PFN is applied on each token vector to generate non-linear hierarchical features. Add and Norm represents residual and layer norm normalization which are included on top of both MHSA and PFN to stay away from vanishing and exploding gradients.

\subsubsection{Self-Attention (SA)}   SA is a much better alternative compared to convolution and recurrent layers to encode global contextual information. For a sequence of input tokens, SA updates each input token vector by encoding global contextual information i.e., it expresses each token vector as a weighted sum of all the token vectors where the weights are given by attention scores. The final input representation matrix X is transformed into Query ($Q \in R^{n \: \times \: q}$), Key ($K \in R^{n \: \times \: k}$) and Value ($V \in R^{n \: \times \: v}$) matrices using three weight matrices $W^{Q} \in R^{e \: \times \: q}$, $W^K \in R^{e \: \times \: k}$  and $W^V \in R^{e \: \times \: v}$.  Here $q=k=v=  \frac{e}{h}$. Here h represents the number of self-attention heads. The output of SA layer is computed as 

\begin{enumerate}
    \item  Compute similarity matrix ( $S \in R^{n \: \times \: n}$) as $Q.K^T$.
    \item To obtain stable gradients, scale the similarity matrix values using $\sqrt{q}$ and then use softmax to convert similarity scores to probability values to get matrix $P \in R^{n \: \times \: n}$. Formally,  $P=Softmax((Q.K^T)/\sqrt{q})$
   \item  Compute the final weighted values matrix $Z \in R^{n \: \times \: v}$ as $P.V$

\end{enumerate}

\subsubsection{Multi-Head Self Attention (MHSA)} With only one self-attention layer, the meaning of a word may largely depend on the same word itself. To avoid this, SA is applied multiple times in parallel each with different weight matrices. Thus, MHSA allows the transformer to attend to multiple positions while encoding a word. Let $Z_1$, $Z_2$, $Z_3$,..,$Z_h$ represent the weighted values matrices of h self-attention heads. Then the final weighted value matrix is obtained by concatenating all these individual weight matrices and then projecting it.
\begin{equation}
    MHSA(X)= [Z_1,Z_2,Z_3,…,Z_h ].W^O
\end{equation}
Where $MHSA(X) \in R^{n \: \times \: e}$, $W^O \in R^{hv \: \times \: e}$  and $[Z_1,Z_2,Z_3,…,Z_h ] \in R^{n \: \times \: hv}$

\subsubsection{Position-wise Feed Forward Network (PFN)} Two linear layers with a non-linear activation constitutes the PFN. PFN is applied to every input token vector. Models like BERT uses Gelu \cite{hendrycks2016gaussian} activation function. Here the parameters of PFNs applied on each of the token vectors are the same. Formally, 
\begin{equation}
    PFN(y)=Gelu(yW_1+ b_1 ) W_2+ b_2
\end{equation}

\subsubsection{Add and Norm} Add represents residual connection while Norm represents layer normalization. Add and Norm is applied on both MHSA and PFN of transformer encoder to stay away from vanishing and exploding gradients.

In general, a transformed-based PLM consists of a sequence of transformer encoder layers after the embedding layer. Each transformer encoder layer updates the input token vectors by encoding global contextual information. By updating the input token vector using a sequence of transformer encoders help the model to encode more language information. Formally,
\begin{equation}
    \hat{E}_{m-1}=LN( E_{m-1}+MHSA(E_{m-1}))
\end{equation}
\begin{equation}
    E_m= LN(\hat{E}_{m-1}+PFN(\hat{E}_{m-1}))
\end{equation}

Here LN represents Layer Normalization,  $\hat{E}_{m-1}$ represents the output after applying Add and Norm over the output of MHSA and $E_m$ represents the output after applying Add and Norm over the output of PFN in $m^{th}$ encoder layer. Overall,  $E_m$ represents the output of $m^{th}$ encoder layer with $E_{m-1}$ as input. Here the input for the first transformer encoder layer is, $E_0=X$.

\subsection{Self-Supervised Learning}
Deep learning algorithms dominated rule-based and machine learning algorithms in the last decade. This is because deep learning models can learn features automatically which eliminates the requirement of expensive feature engineering and process the inputs in an end-to-end manner i.e., take raw inputs and give the decisions.  The success of the deep learning algorithms comes from the knowledge gained during training from human-labeled instances. However, supervised learning has a lot of limitations a) with less data, the model may get overfitted and prone to bias b) some of the domains like biomedical are supervision starved i.e., difficult to get labeled data. In general, we expect models to be close to human intelligence i.e., more general and make decisions with just a few samples. This desire of developing models with more generalization ability and learning from fewer samples has made the researchers focus on other learning paradigms like Self-Supervised Learning \cite{liu2021self}. 

\begin{figure}[t]
\begin{center}
\includegraphics[width=7cm, height=2cm]{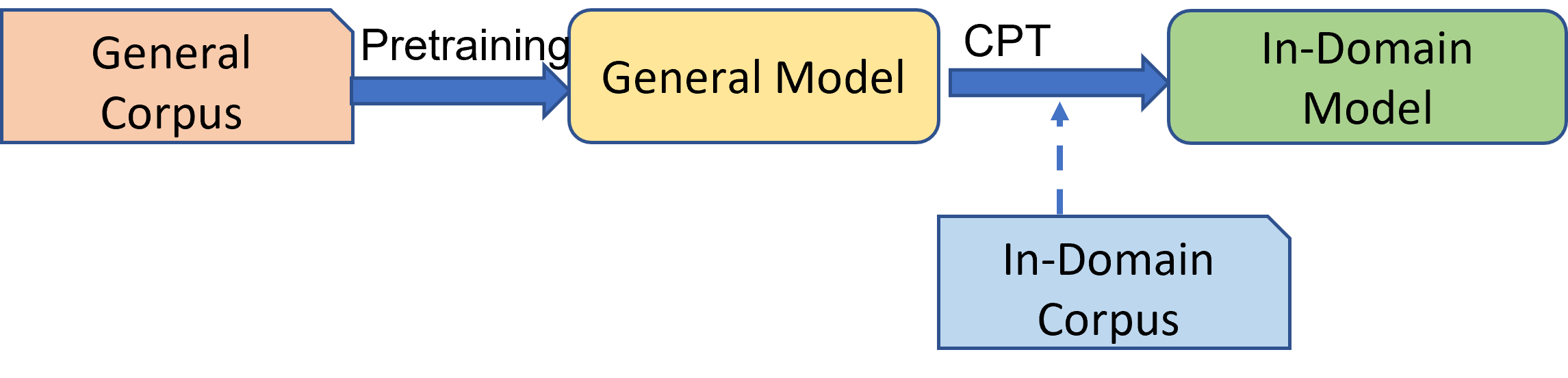}
\caption{\label{ammu-cpt} Continual Pretraining (CPT) } 
\end{center}
\end{figure}

Robotics is the first AI field to use self-supervised learning methods \cite{liu2021self}. Over the last five years, self-supervised learning has become popular in other AI fields like natural language processing \cite{qiu2020pre,liu2020survey,kalyan2021ammus}, computer vision \cite{khan2021transformers,han2020survey}, and speech processing \cite{baevski2020wav2vec,sivaraman2020self}. SSL is a new learning paradigm that draws inspiration from both supervised and unsupervised learning methods. SSL is similar to unsupervised learning as it does not depend on human-labeled instances. It is also similar to supervised learning as it learns using supervision. However, in SSL the supervision is provided by the pseudo labels which are generated automatically from the pretraining data. SSL involves pretraining the model over a large unlabelled corpus using one or more pretraining tasks. The pseudo labels are generated depending on the definitions of pre-training tasks. SSL methods fall into three categories namely Generative, Contrastive, and Generate-Contrastive \cite{liu2021self}. In Generative SSL, encoder maps input vector x to vector y, and decoder recovers x from y (e.g., masked language modeling). In Contrastive SSL, the encoder maps input vector x to vector y to measure similarity (e.g., mutual information maximization). In Generate-Contrastive SSL, fake samples are generated using encoder-decoder while the discriminator identifies the fake samples (e.g., replaced token detection).  For more details about different SSL methods, please refer to the survey paper written by Liu et al. \cite{liu2021self}.

\section{T-BPLMs Core Concepts}
\label{core-concepts-sec}

\subsection{Pretraining Methods}
\begin{figure*}[h]
\begin{center}
\includegraphics[width=12cm, height=5cm]{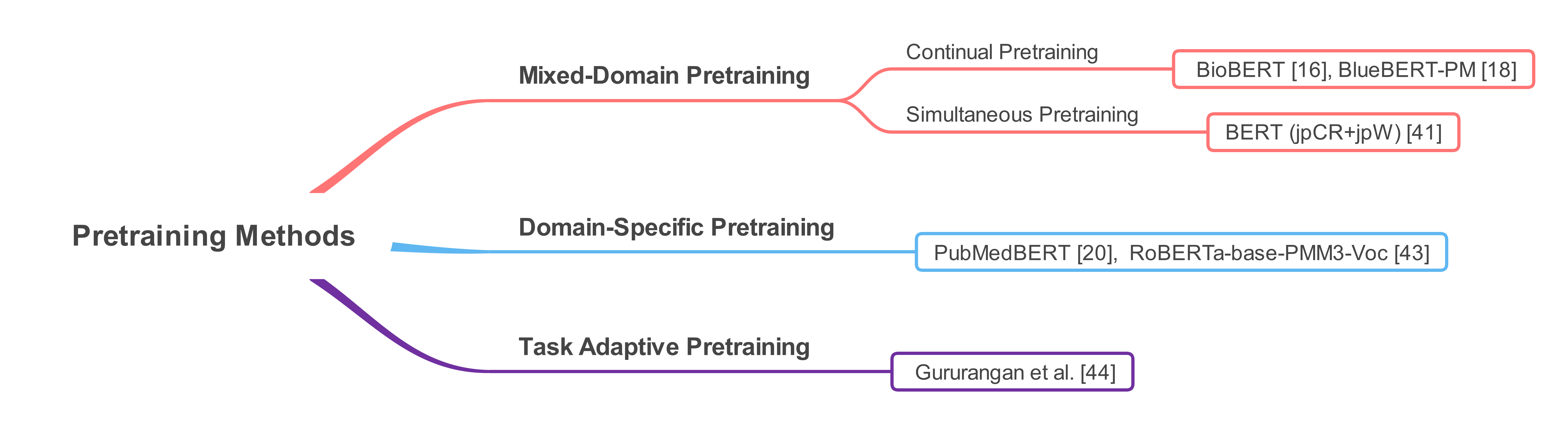}
\caption{\label{ammu-pretraining-methods} Pretraining Methods } 
\end{center}
\end{figure*}

SSL involves pretraining on large volumes of unlabeled data using one or more tasks. Pretraining allows the model to learn language representations that are useful across tasks. Moreover, pretraining gives the model a better initialization which avoids training from scratch and overfitting in low data situations. Pretraining methods in biomedical NLP fall into three categories as shown in Figure \ref{ammu-pretraining-methods}. 

\begin{figure}[h]
\begin{center}
\includegraphics[width=7cm, height=1.2cm]{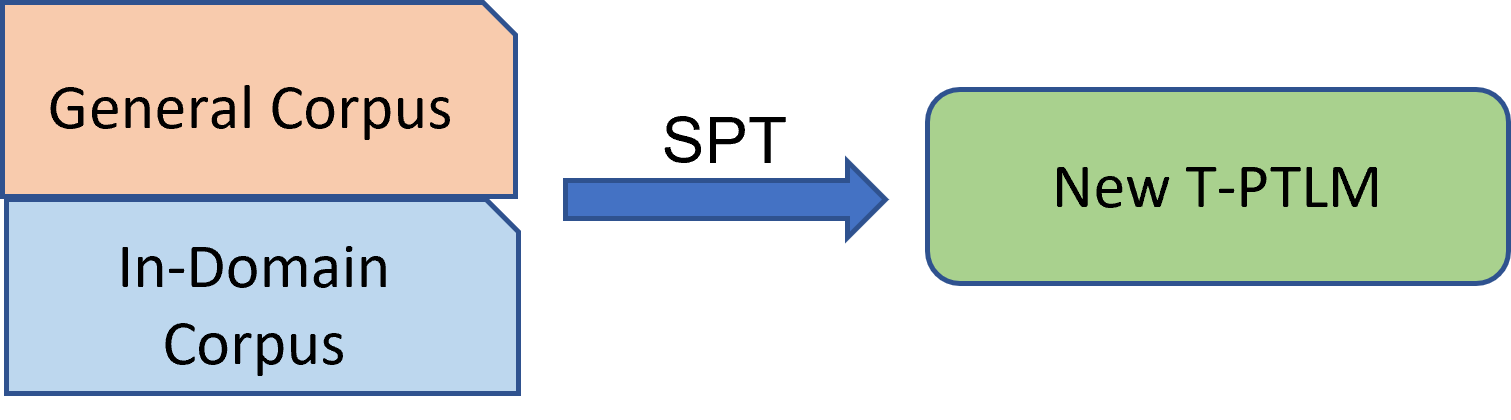}
\caption{\label{ammu-spt} Simultaneous Pretraining (SPT) } 
\end{center}
\end{figure}

\subsubsection{Mixed-Domain Pretraining (MDPT)} Mixed domain pretraining involves training the model using both general and in-domain text. Depending on whether the pretraining is done simultaneously or not, mixed domain pretraining can be classified into a) Continual pretraining – initially the model is pre-trained over general domain text and then adapted to the biomedical domain \cite{lee2020biobert} and b) Simultaneous pretraining – the model is pre-trained over the combined corpora having both general and in-domain text where the in-domain text is up sampled to ensure balanced pretraining \cite{wada2020pre}.

\textbf{Continual Pretraining (CPT)} : It is the standard approach followed by the biomedical NLP research community to develop transformer-based BPLMs. It is also referred to as further pretraining. In this approach, the model is initialized with general PLM weights and then the model is adapted to in-domain by further pretraining on large volumes of in-domain text (refer Figure \ref{ammu-cpt}). For example, BioBERT is initialized with general BERT weights and then further pretrained on PubMed abstracts and PMC full-text articles \cite{lee2020biobert}. In the case of BlueBERT, the authors used both PubMed abstracts and MIMIC-III clinical notes for continual pretraining \cite{peng2019transfer}. 

\begin{figure}[t!]
\begin{center}
\includegraphics[width=7cm, height=1cm]{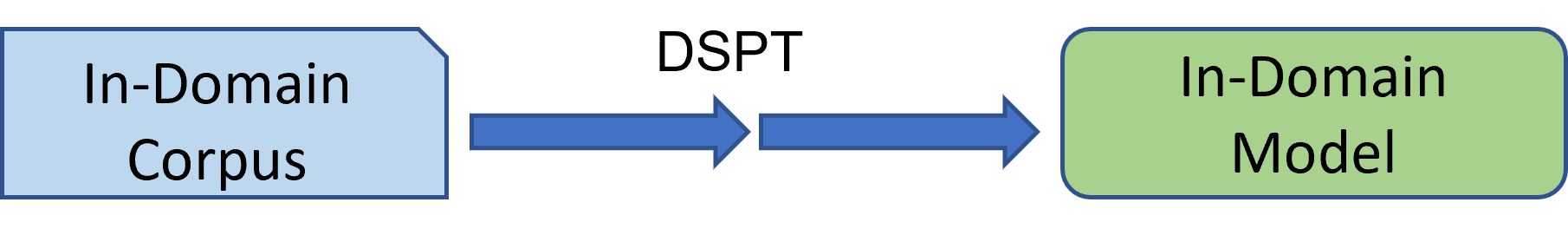}
\caption{\label{ammu-dspt} Domain-Specific Pretraining (DSPT) } 
\end{center}
\end{figure}

\textbf{Simultaneous Pretraining (SPT)} : Continual pretraining achieved good results by adapting general models to the biomedical domain \cite{lee2020biobert,yang2020clinical,huang2020clinical,si2019enhancing,peng2019transfer}. However, it requires large volumes of in-domain text. Otherwise, CPT may result in suboptimal performance. Simultaneous pretraining comes to the rescue when only a small amount of in-domain text is available. Here, the pretraining corpora consist of both in-domain and general domain text where the in-domain text is up sampled to ensure a balanced pretraining (refer Figure \ref{ammu-spt}). For example, BERT (jpCR+jpW) \cite{antoun2020arabert} is developed by simultaneous pretraining over a small amount of Japanese clinical text and a large amount of Japanese Wikipedia text. This model outperformed UTH-BERT in clinical text classification. UTH-BERT \cite{kawazoe2020clinical} is trained from scratch over Japanese clinical text.

\subsubsection{Domain-Specific Pretraining (DSPT)} The main drawback in continual pretraining is the general domain vocabulary. For example, the WordPiece vocabulary in BERT is learned over English Wikipedia and Books Corpus \cite{devlin2019bert}. As a result, the vocabulary does not represent the biomedical domain and hence many of the biomedical words are split into several subwords which hinders the model learning during pretraining and fine-tuning. Moreover, the length of the input sequence also increases as many of the in-domain words are split into several subwords. DSPT over in-domain text allows the model to have the in-domain vocabulary (refer Figure \ref{ammu-dspt}). For example, PubMedBERT is trained from scratch using PubMed abstracts and PMC full-text articles \cite{gu2020domain}. PubMed achieved state-of-the-art results in the BLURB benchmark. Similarly, RoBERTa-base-PM-M3-Voc is trained from scratch over PubMed and PMC and MIMIC-III clinical notes \cite{lewis2020pretrained}.

\subsubsection{Task Adaptive Pretraining (TAPT)} 

\begin{figure}[h]
\begin{center}
\includegraphics[width=7cm, height=1.6cm]{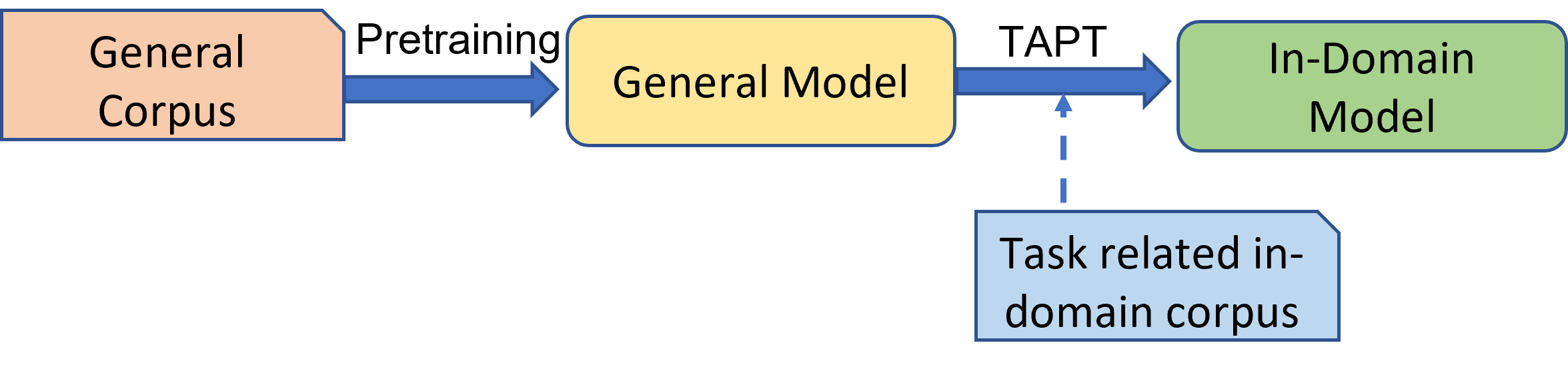}
\caption{\label{ammu-tapt} Task Adaptive Pretraining (TAPT) } 
\end{center}
\end{figure}

Both DSPT and MDPT require training the model over large volumes of text to allow the model to learn domain-specific knowledge which helps to perform it to perform better in downstream tasks. Pretraining over in-domain text allows the model to learn universal in-domain representations which are useful to all the in-domain tasks. However, pretraining over large volumes of text is expensive in terms of both computational resources and time. Task Adaptive Pretraining (TAPT) is based on the hypothesis that pretraining over task-related unlabelled text allows the model to learn both domain and task-specific knowledge \cite{gururangan2020don} (refer Figure \ref{ammu-tapt}). In TAPT, task-related unlabelled sentences are gathered, and then the model is further pretrained. TAPT is less expensive compared to other pretraining methods as it involves pretraining the model over a relatively small corpus of task-related unlabelled sentences. 

\subsection{Pretraining Tasks}
\begin{figure*}[h]
\begin{center}
\includegraphics[width=12cm, height=5cm]{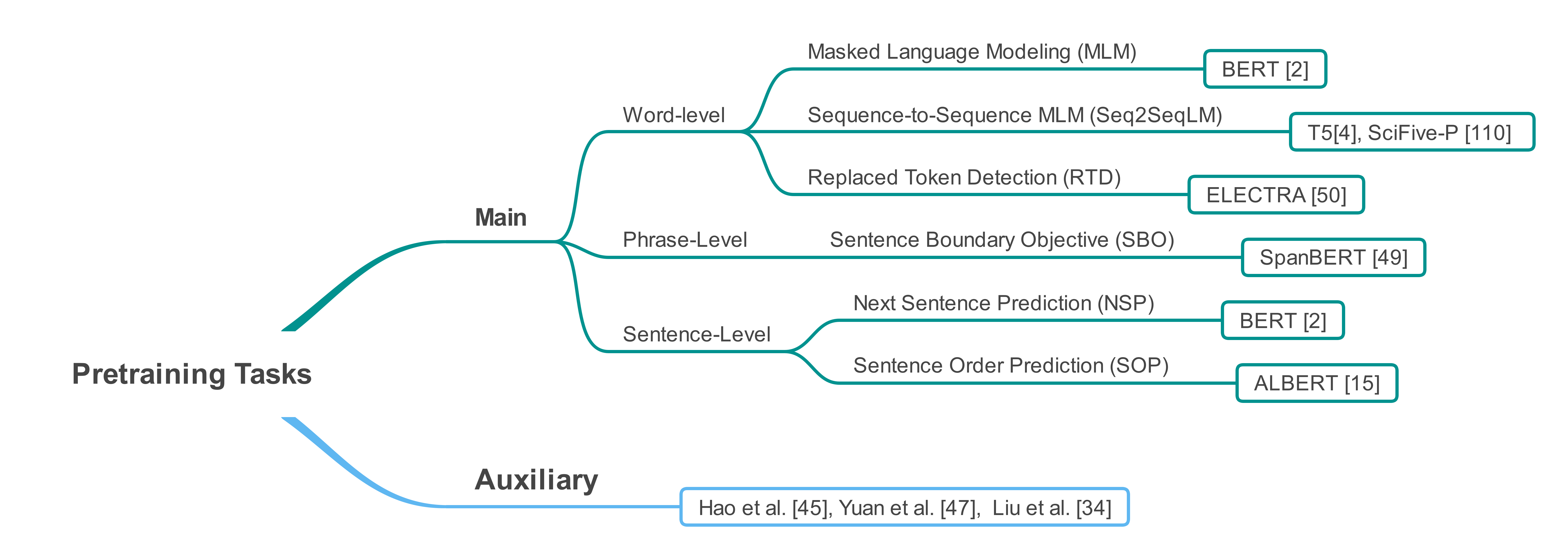}
\caption{\label{ammu-pretraining-tasks} Pretraining Tasks } 
\end{center}
\end{figure*}

During pretraining, the language models learn language representations based on the supervision provided by one or more pretraining tasks. A pretraining task is a pseudo-supervised task whose labels are generated automatically. A pretraining task can be main or auxiliary. The main pretraining tasks allow the model to learn language representations while auxiliary pretraining tasks allow the model to gain knowledge from human-curated sources like Ontology \cite{hao2020enhancing,michalopoulos2021umlsbert,yuan2020coder,liu2021self}. The classification of pretraining tasks is given in Figure \ref{ammu-pretraining-tasks} and a brief summary of various pretraining tasks is presented in Table \ref{table-pretraining-tasks}.   

\begin{table*}[t!]
\begin{center}
{\renewcommand{\arraystretch}{1.5}% for the vertical padding
\begin{tabular}{|p{2cm}|p{2cm}|p{9.5cm}|p{2.5cm}|}
\hline
Pretraining Task  & Type         & Key points    & Models    \\ \hline

MLM                        & word-level     & Model learns by predicting the masked tokens. Less   training signal per instance as the model predicts only 15\% of the tokens.  & BERT \cite{devlin2019bert}                             \\ \hline

MLM + Dynamic Masking      & word-level     & Dynamic Masking allows masking different tokens in   the sentences for different epochs due to which the model learns more by   predicting different tokens every time.  & RoBERTa \cite{liu2019roberta}   \\ \hline

MLM + Whole Word Masking   & word-level     & Whole word masking is more challenging as it is   difficult to predict the entire word compared to a subword.   & PubMedBERT \cite{gu2020domain} \\ \hline

MLM + Whole Entity Masking & word-level     & Whole entity masking allows the model to learn   entity-centric knowledge.    & MC-BERT \cite{zhang2020conceptualized}   \\ \hline

MLM + Whole Span Masking   & word-level     & Whole span masking allows the model to learn more   linguistic knowledge.     & MC-BERT \cite{zhang2020conceptualized}   \\ \hline

NSP    & sentence-level & Allows the model to learn sentence-level reasoning   skills which are useful in tasks like NLI. Less challenging as it involves   topic prediction which is a relatively easy task. & BERT \cite{devlin2019bert}    \\ \hline

SOP    & sentence-level & Allows the model to learn sentence-level reasoning   skills by modeling inter-sentence coherence. More challenging compared to NSP   as SOP involves only sentence coherence.       & ALBERT \cite{lan2019albert}     \\ \hline

SBO   & phrase-level   & Model predicts the masked tokens in a span based on   boundary token representations and position embeddings.    & SpanBERT \cite{joshi2020spanbert}  \\ \hline

RTD   & word-level     & Model checks every token whether it is replaced or   not. More efficient compared to MLM as it involves all the tokens in the   input.  & ELECTRA \cite{clark2019electra}   \\ \hline
 
\end{tabular}}
\end{center}
\caption{\label{table-pretraining-tasks} Summary of pretraining tasks.} 
\end{table*}

\subsubsection{Main Pretraining Tasks} 
The main pretraining tasks allow the model to learn language representations. Some of the commonly used main pretraining tasks are masked language modelling (MLM) \cite{devlin2019bert}, replaced token detection (RTD) \cite{clark2019electra}, sentence boundary objective (SBO) \cite{joshi2020spanbert}, next sentence prediction (NSP)  \cite{devlin2019bert} and sentence order prediction (SOP) \cite{lan2019albert}.

\textbf{Masked Language Modeling (MLM)}.  It is an improved version of Language Modeling which utilizes both left and right contexts to predict the missing tokens \cite{devlin2019bert}. The main drawback in Unidirectional LM (Forward LM or Backward LM) is the inability to utilize both left and right contexts at the same time to predict the tokens. However, the meaning of a word depends on both the left and right contexts. Devlin et al. \cite{devlin2019bert} utilized MLM as a pretraining task for learning the parameters of the BERT model. Formally, for a given sequence $x$ with tokens $\{x_1, x_2, …, x_m\}$, a subset of tokens is randomly chosen and these tokens are replaced.  The authors replaced tokens, 80\% of the time with a special token ‘[MASK]’, 10\% of the time with a random token, and 10\% of the time with the same token. This is done to handle the mismatch between pretraining and fine-tuning phases. Formally,

\begin{equation}
    L_{MLM} = - \frac{1}{|m(x)|}\sum_{i \in m(x)} logP(x_i/\hat{x})
\end{equation}
where $\hat{x}$ is the masked version of $x$ and $m(x)$ represents the set of masked token positions.

Some of the improvements like dynamic masking \cite{liu2019roberta}, whole word masking \cite{devlin2019bert,gu2020domain,cui2019pre}, whole entity masking \cite{pergola2021boosting,zhang2020conceptualized}, and whole span masking \cite{zhang2020conceptualized} are introduced in MLM to further improve its efficiency as a pretraining task. Delvin et al. \cite{devlin2019bert} used static masking to replace the tokens i.e., the input sentences are masked once during pre-processing and the model predicts the same masked tokens in the input sentences for every epoch during pretraining.  In the case of dynamic masking \cite{liu2019roberta}, different tokens are masked in the input sentence for different epochs which prevents the model from predicting the same masked tokens in every epoch and hence it learns more.  Whole word masking is much more challenging as the model has to predict the entire word rather than part of a word. In the case of the whole entity and span maskings, in-domain entities and phrases in the input sentences are identified and then masked rather than masking the randomly chosen tokens. As a result, the model learns entity-centric and in-domain linguistic knowledge during pretraining which enhances the performance of the model in downstream tasks \cite{pergola2021boosting,zhang2020conceptualized}. For example, Zhang et al. \cite{zhang2020conceptualized} trained MC-BERT using NSP and MLM with whole entity and span maskings. Michalopoulos et al. \cite{michalopoulos2021umlsbert} used novel multi-label loss-based MLM along with NSP to further pretrain ClinicalBERT on MIMIC-III clinical notes to get UmlsBERT. Novel multilabel loss-based MLM allows the model to connect all the words under the same concept. Sequence-to-Sequence MLM (Seq2SeqLM) is an extension of MLM to models based on encoder-decoder architecture. Models like T5 \cite{raffel2020exploring} are pretrained using  Seq2SeqLM pretraining task.

\textbf{Replaced Token Detection (RTD) \cite{clark2019electra}}.  It is a novel pretraining task that involves verifying whether each token in the input is replaced or not.  Initially, some of the tokens in the input sentences are replaced with words predicted by a small generator network, and then the model (discriminator) is asked to predict the status of each word as replaced or not.  The two advantages of RTD over MLM are a) RTD provides more training signal compared to MLM as RTD involves checking the status of every token in the input rather than a subset of randomly chosen tokens like MLM. and b) Unlike MLM, RTD does not use any special tokens like ‘[MASK]’ to corrupt the input. So, it avoids the mismatch problem that the special token ‘[MASK]’ is seen only during pretraining but not during fine-tuning. Formally, 

\begin{equation}
    L_{RTD} = -\frac{1}{|\hat{x}|} \sum_{i=1}^{|\hat{x}|} logP(t/\hat{x_i})
\end{equation}
where $\hat{x}$ is the corrupted version of $x$ and $t=1$ when the token is not a replaced one.

\textbf{Span Boundary Objective (SBO) \cite{joshi2020spanbert}}.  It is a novel pretraining task that involves predicting the entire masked span based on the context. Initially, a contiguous span of tokens is randomly chosen and masked and then the model is asked to predict the masked tokens in the span based on the token representations at the boundary. In the case of MLM, the model predicts the masked token based on the final hidden vector of the masked token. However, in the case of SBO, the model predicts the masked token in the span based on the final hidden vectors of the boundary tokens and the position embedding of the masked token. SBO is more challenging as it is difficult to predict the entire span \textit{“frequent bathroom runs”} than predicting \textit{“frequent”} when the model already sees \textit{“bathroom runs”}. SBO helps the model to achieve better results in span extraction-based tasks like entity extraction and question answering \cite{portelli2021bert,joshi2020spanbert}. Let $s$ and $e$ represent the start and end indices of the span in the input sequence. Then, each token $x_i$ in the span is predicted based on the final hidden vectors of the boundary tokens $x_{s-1}$, $x_{e+1}$ and its position embedding $p_{i-s+1}$. Then

\begin{equation}
    L_{SBO} = -\frac{1}{|S|} \sum_{i\in S} logP(x_i/y_i)
\end{equation}
where $y_i=g(x_{s-1},x_{e+1},p_{i-s+1}) $, $g()$ represents feed-forward network of two layers and $S$  represents the positions of tokens in contiguous span.

\textbf{Next Sentence Prediction (NSP) \cite{devlin2019bert}}.  NSP is a sentence-level pretraining task that involves predicting whether given two sentences appear consecutively or not . It is basically a two-way sentence pair classification task. Formally, for a given sentence pair $(x, y)$, the model has to predict one of the two labels $\{IsNext, IsNotNext\}$ depending on whether the two sentences are consecutive or not. NSP helps the model to learn sentence-level reasoning skills which are useful in downstream tasks involving sentence pairs like natural language inference, text similarity, and question answering \cite{devlin2019bert}. For a balanced pretraining, the training examples are chosen in a 1:1 ratio i.e., 50\% are positive and the rest negative.  Let $z$ represents aggregate vector representation of the sentence pair (x, y). Then,

\begin{equation}
    L_{NSP} = -logP(t/z)
\end{equation}
where $t=1$ when the two sentences $x$ and $y$ are consecutive.

\textbf{Sentence Order Prediction (SOP) \cite{lan2019albert}}. SOP is a novel sentence-level pretraining task which models inter-sentence coherence. Like NSP, SOP is a two-way sentence pair classification. Formally, for a given sentence pair $(x, y)$, the model has to predict one of the two labels $\{IsSwapped, IsNotSwapped\}$ depending on whether the sentences are swapped or not. For a balanced pretraining, the training examples are chosen in a 1:1 ratio i.e., 50\% are swapped and the rest are not swapped.  Unlike NSP which involves the prediction of both topic and coherence, SOP involves only sentence coherence prediction \cite{lan2019albert}. Topic prediction is comparatively easier which questions the effectiveness of NSP as a pretraining task \cite{liu2019roberta,joshi2020spanbert,lan2019albert}. Let $z$ represent aggregate vector representation of the sentence pair $(x, y)$. Then,
\begin{equation}
    L_{SOP} = -logP(t/z)
\end{equation}
where $t=1$ when the two sentences $x$ and $y$ are not swapped.

\subsubsection{Auxiliary Pretraining Tasks}
Auxiliary pretraining tasks help to inject knowledge from human-curated sources like UMLS \cite{bodenreider2004unified} into in-domain models to further enhance them. For example, the triple classification pretraining task involves identifying whether two concepts are connected by the relation or not \cite{hao2020enhancing}. This auxiliary task is used by Hao et al. \cite{hao2020enhancing} to inject UMLS relation knowledge into in-domain models. Yuan et al. \cite{yuan2020coder} used two auxiliary pretraining tasks based on multi-similarity Loss and Knowledge embedding loss to further pretrain BioBERT on UMLS. Similarly, Liu et al. \cite{liu2021self} used multi-similarity loss-based pretraining task to inject UMLS synonym knowledge into PubMedBERT. 

\subsection{Fine-Tuning Methods}
Pretraining allows the model to learn general or in-domain knowledge which is useful across the tasks. However, for a model to perform well in a particular task, it must have task-specific knowledge along with general or in-domain knowledge. The model gains task-specific knowledge by fine-tuning on the task-specific datasets. Task-specific layers are included on the top of transformer-based BPLMs. For example, to perform text classification, we need a) a contextual encoder to learn contextual token representations from the given input token vectors and b) a classifier to project the final sequence vector and then generate the probability vector. Here classifier is the task-specific layer which is usually a softmax layer in text classification. Fine-tuning methods fall into two categories.

\subsubsection{Intermediate Fine-Tuning (IFT)} IFT on large, related datasets allows the model to learn more domain or task-specific knowledge which improves the performance on small target datasets. IFT can be done in following four ways

\textbf{Same Task Different Domain} – Here, the source and target datasets are from the same task but different domains. Model can be fine-tuned on general domain datasets before fine-tuning on small in-domain datasets  \cite{cengiz2019ku_ai,yang2020measurement,wang2020learning,yoon2019pre}. For example, Cengiz et al. \cite{cengiz2019ku_ai} fine-tuned in-domain model on general NLI datasets like SNLI \cite{bowman2015large} and MNLI \cite{williams2018broad} before fine-tuning on MedNLI \cite{romanov2018lessons}.

\textbf{Same Task Same Domain} – Here, the source and target datasets are from the same task and domain. But the source dataset is a more generic one while the target dataset is more specific \cite{sun2021biomedical,gao2021pre}. For example, Gao et al. \cite{gao2021pre} fine-tuned BlueBERT on a large general biomedical NER corpus like MedMentions \cite{mohan2018medmentions} or Semantic Medline before fine-tuning on the small target NER corpus. 

\textbf{Different Task Same Domain} – Here, the source and target datasets are from different tasks but the same domain. Fine-tuning on source dataset which is from the same domain allows the model to gain more domain-specific knowledge which improves the performance of the model on the same domain target task \cite{mccreery2019domain}.  McCreery et al. \cite{mccreery2019domain} fine-tuned the model on the medical question-answer pairs dataset to enhance its performance on the medical question similarity dataset. 

\textbf{Different Task Different Domain} – Here, the source and target datasets are from different tasks and different domains. For example, Jeong et al. \cite{jeong2020transferability} fine-tuned BioBERT on a general MultiNLI dataset to improve the performance of the model in biomedical QA. Here the model learns sentence level reasoning skills which are useful in biomedical QA.

\subsubsection{Multi-Task Fine-Tuning} Multi-task fine-tuning allows the model to be fine-tuned on multiple tasks simultaneously \cite{liu2019multi,zhang2021survey,khan2020mt}. Here the embedding and transformer encoder layers are common for all the tasks and each task has a separate task-specific layer. Multi-task fine-tuning allows the model to gain domain as well as task-specific reasoning knowledge from multiple tasks. At the same time, due to the increase in training set size, the model is less prone to over-fitting. Multi-task fine-tuning is more useful in low resource scenarios which are common in the biomedical domain \cite{khan2020mt}. Moreover, having a single model for multi-tasks eliminates the need of deploying separate models for each task which saves computational resources, time, and deployment costs \cite{mulyar2020mt}. Multi-task fine-tuning may not provide the best results all the time \cite{mulyar2020mt}. In such cases, multi-task fine tuning can be applied iteratively to identify the best possible subset of tasks \cite{mahajan2020identification}. For example, Mahanjan et al. \cite{mahajan2020identification} applied multi-task fine-tuning iteratively to choose the best subset of related datasets. Finally, the authors fine-tuned the model on best subset of related datasets and achieved the best results on the Clinical STS \cite{wang20202019} dataset.   After multi-task fine-tuning, the model can be further fine-tuned on the target specific dataset separately to further enhance the performance of the model \cite{peng2020empirical}.  

\subsection{Embeddings}
\label{embeddings-ssec}
\begin{figure*}[h]
\begin{center}
\includegraphics[width=12cm, height=5cm]{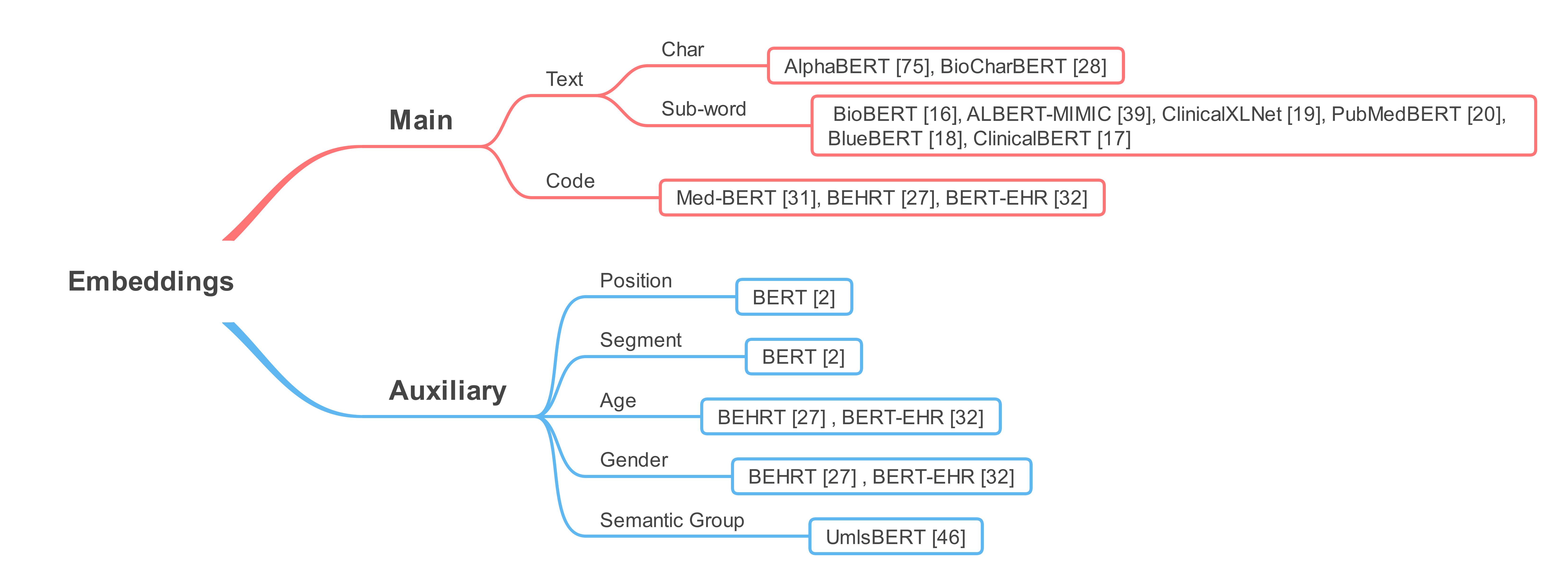}
\caption{\label{ammu-embeddings} Embeddings in T-BPLMs } 
\end{center}
\end{figure*}

Embeddings represent the data in a low-dimensional space. Embeddings in transformer-based BPLMs fall into two categories namely main and auxiliary. The main embeddings map the given input sequence to a sequence of vectors while auxiliary embeddings provide additional useful information. Figure \ref{ammu-embeddings} shows the classification of embeddings.

\subsubsection{Main Embeddings} Text embeddings map the given sequence of words into a sequence of vectors. Text embeddings can be char, subword or code-based. 

\textbf{Character Embeddings} -  In character embeddings, the vocabulary consists of letters, punctuation symbols, special characters and numbers only. Each character is represented using an embedding. These embeddings are initialized randomly and learned during model pretraining. ELMo embedding model uses CharCNN to generate word representations from character embeddings \cite{peters2018deep}. Inspired by ELMo, BioCharBERT also uses CharCNN on the top of character embeddings to generate word representations \cite{el2020characterbert}. AlphaBERT \cite{chen2020modified} also uses character embeddings. Unlike CharacterBERT, AlphaBERT directly combines character embeddings with position embeddings and then applies a stack of transformer encoders. The main advantage with character embeddings is the small size of vocabulary as it includes only characters. The disadvantage is longer pretraining times \cite{el2020characterbert}. As the sequence length increases with character level embeddings, models are slow to pre-train. 

\textbf{Subword Embeddings}- In subword embeddings, the vocabulary consists of characters and the most frequent subwords and words. The main principle driving the subword embedding vocabulary construction is that frequent words should be represented as a single word and rare words should be represented in terms of meaningful subwords. Subword embedding vocabularies are always moderate in size as they use sub-words to represent rare and misspelled words. Some of the popular algorithms to generate vocabulary for sub-word embeddings are Byte-Pair Encoding (BPE) \cite{sennrich2016neural}, Byte-Level BPE \cite{radford2019language}, Word-Piece \cite{wu2016google}, Unigram \cite{kudo2018subword}, and Sentencepiece \cite{kudo2018sentencepiece}. 

\textit{Byte-Pair Encoding (BPE) \cite{sennrich2016neural}} - It starts with a base vocabulary having all the unique characters in the training corpus. It augments the base vocabulary with the most frequent pairs until the desired vocabulary size is achieved. Byte-Pair Encoding algorithm can be summarized as
\begin{enumerate}
    \item Prepare a large training corpus and fix the vocabulary size.
    \item Generate a base vocabulary having all the unique characters in the training corpus.
    \item Calculate the frequency of all the words in the corpus.
    \item Augment the vocabulary with the most frequently occurring pair.
    \item Until the desired vocabulary size is achieved, repeat step 4.  
\end{enumerate}

\textit{Byte-Level BPE \cite{radford2019language}} - In Byte-Level BPE, each character is represented as a byte, and the rest of the procedure is the same as in BPE. Text is converted into a sequence of bytes and the most frequent byte pair is added into the base vocabulary until the desired size is achieved. Byte-Level BPE is extremely beneficial in the multilingual scenario. GPT-2 \cite{radford2019language} and RoBERTa \cite{liu2019roberta} use Byte-Level BPE embeddings.

 \textit{WordPiece \cite{wu2016google}} - The working of WordPiece is almost the same as BPE. Word-Piece and BPE differ in the strategy used in selecting the symbol pair to augment the base vocabulary. BPE chooses the most frequent symbol pair while Word-Piece uses a language model to choose the symbol pair. BERT \cite{devlin2019bert}, DistilBERT \cite{sanh2019distilbert}, and ELECTRA model use WordPiece embeddings.
  
 \textit{SentencePiece \cite{kudo2018sentencepiece}} -  A common problem in BPE and WordPiece is that they assume that words in the input sentences are separated by space. However, this assumption is not applicable in all languages. To overcome this, SentencePiece treats space as a character and includes it in the base vocabulary. The final vocabulary is generated iteratively using BPE or Unigram.  XLNet \cite{yang2019xlnet}, ALBERT \cite{lan2019albert}, and T5 \cite{raffel2020exploring} models use SentencePiece embeddings.
  
 \textit{Unigram \cite{kudo2018subword}} - Unigram is similar to BPE and WordPiece in fixing the vocabulary size at the beginning itself. BPE and Word-piece start with a small base vocabulary and then augments the base vocabulary for a certain number of iterations. Unlike BPE and Word-Piece, Unigram starts with a large base vocabulary and then iteratively trims the symbols to arrive at a small final vocabulary. It is not used directly in any of the models. SentencePiece uses the Unigram algorithm to generate the final vocabulary. 
 
 \begin{figure*}[h]
\begin{center}
\includegraphics[width=18cm, height=12cm]{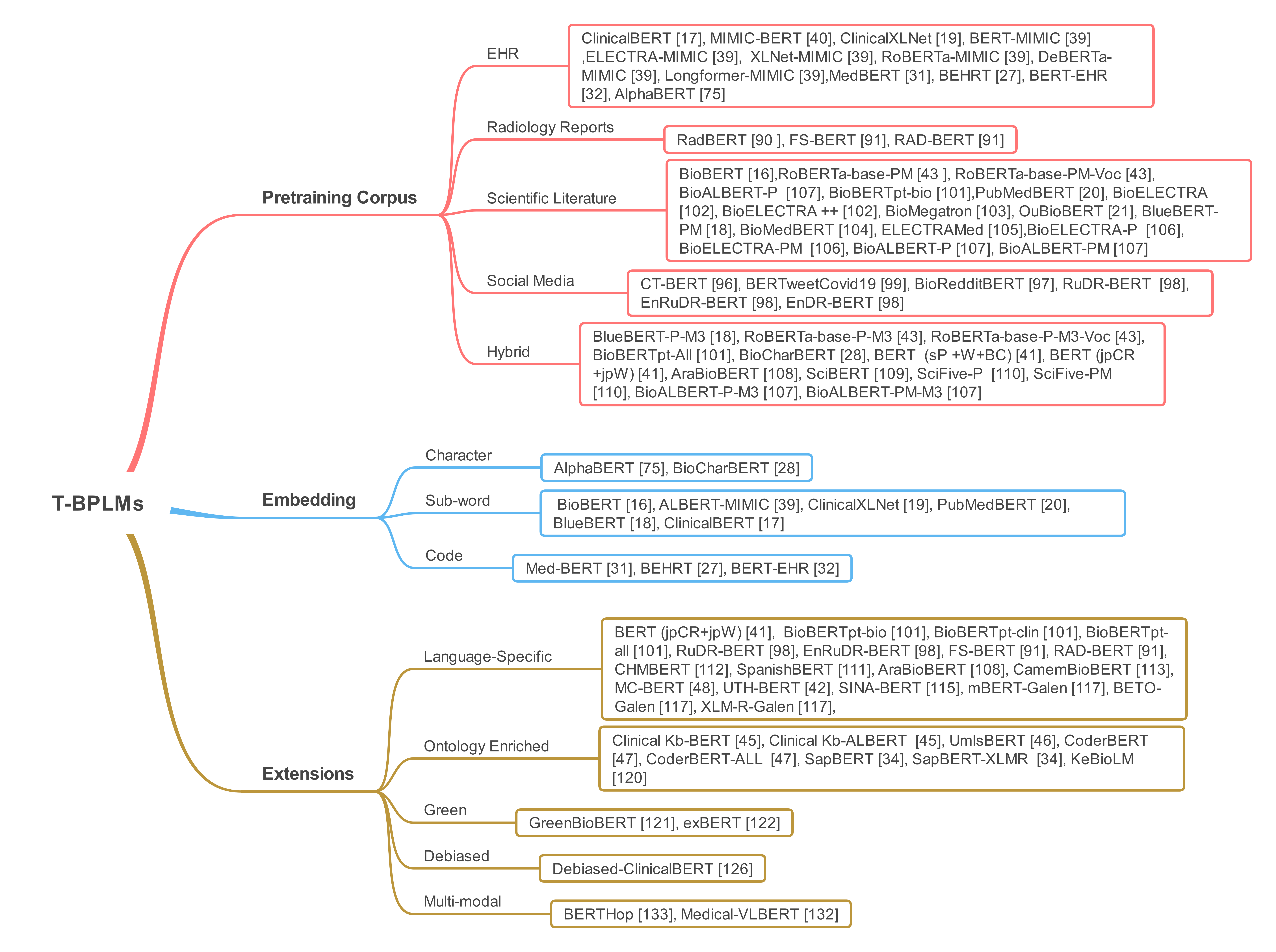}
\caption{\label{ammu-taxonomy} T-BPLMs taxonomy } 
\end{center}
\end{figure*}
 
 \textbf{Code embeddings} - Code embeddings map the given sequence of codes into a sequence of vectors. For example, in the case of models like BERT-EHR \cite{meng2021bidirectional}, MedBERT \cite{rasmy2021med}, and BEHRT \cite{li2020behrt}, the input is not a sequence of words. Instead, input is patient visits. Each patient visit is represented as a sequence of codes. The number of code embeddings varies from model to model. For example, MedBERT and BEHRT include embeddings only for disease codes while BERT-EHR includes embeddings for disease, medication, procedure, and clinical notes. 
 
\subsubsection{Auxiliary Embeddings} Main embeddings represent the given input sequence in low dimensional space. The purpose of auxiliary embeddings is to provide additional information to the model so that the model can learn better. For each input token, a representation vector is obtained by summing the main and two or more auxiliary embeddings. The various auxiliary embeddings are 

\textbf{Position Embeddings} - Position embeddings enhance the final input representation of a token by providing its position information in the input sequence. As there is no convolution or recurrence layers which can learn the order of input tokens automatically, we need to explicitly provide the location of each token in the input sequence through position embeddings. Position embeddings can be pre-determined \cite{li2020behrt,meng2021bidirectional} or learned  during model pretraining \cite{devlin2019bert}.  

\textbf{Segment Embeddings} - Segment embeddings help to distinguish tokens of different input sequences. Segment embedding is the same for all the tokens in the same input sequence. 

\textbf{Age Embeddings} - In models like BEHRT \cite{li2020behrt} and BERT-EHR \cite{meng2021bidirectional}, age embeddings are used in addition to other embeddings. Age embeddings provide the age of the patient and help the model to leverage temporal information. Age embedding is the same for all the codes in a single patient visit. 
    
\textbf{Gender Embeddings} - In models like BEHRT \cite{li2020behrt} and BERT-EHR \cite{meng2021bidirectional}, gender embeddings are used in addition to other embeddings. Gender embeddings provide the gender information of the patient to the model. Gender embedding is the same for all the codes in all the patient visits.
    
\textbf{Semantic Group Embeddings} - Semantic group embeddings are used in UmlsBERT \cite{michalopoulos2021umlsbert} to explicitly inform the model to learn similar representations for words from the same semantic group i.e., semantic group embedding is same for all the words which fall into the same semantic group. Besides, it also helps to provide better representations for rare words. 

\begin{table*}[t!]
\begin{center}
{\renewcommand{\arraystretch}{1.5}% for the vertical padding
\begin{tabular}{|p{2.5cm}|p{2cm}|p{1.5cm}|p{3.5cm}|p{1.5cm}|p{4cm}|}
\hline
    \textbf{Model}                              & \textbf{Type} & \textbf{Pretrained from}    & \textbf{Corpus}                        & \begin{tabular}[c]{@{}l@{}}\textbf{Publicly}\\    \\ \textbf{Available}\end{tabular} & \textbf{Evaluation}                                            \\ \hline

ClinicalBERT \cite{alsentzer2019publicly}             & EHR  & BioBERT            & MIMIC-III Clinical Notes      & Yes               & MedNLI and  Clinical Concept Extraction               \\ \hline

ClinicalBERT (discharge) \cite{alsentzer2019publicly}  & EHR  & BioBERT            & MIMIC-III Discharge summaries & Yes              & MedNLI and  Clinical Concept Extraction               \\ \hline

MIMIC-BERT \cite{si2019enhancing}                & EHR  & General BERT       & MIMIC-III Clinical Notes      & Yes                & Clinical Concept Extraction                           \\ \hline

ClinicalXLNet (nursing) \cite{huang2020clinical}   & EHR  & General XLNet      & MIMIC-III Nursing notes       & Yes              & Prolonged Mechanical Ventilation   Prediction problem \\ \hline

ClinicalXLNet (discharge) \cite{huang2020clinical}  & EHR  & General XLNet      & MIMIC-III Discharge notes     & Yes               & Prolonged Mechanical Ventilation   Prediction problem \\ \hline

BERT-MIMIC \cite{yang2020clinical}                & EHR  & General BERT       & MIMIC-III Clinical Notes      & Yes              & Clinical Concept Extraction                           \\ \hline

ELECTRA-MIMIC \cite{yang2020clinical}              & EHR  & General ELECTRA    & MIMIC-III Clinical Notes      & Yes               & Clinical Concept Extraction                           \\ \hline

XLNet-MIMIC \cite{yang2020clinical}              & EHR  & General XLNet      & MIMIC-III Clinical Notes      & Yes             & Clinical Concept Extraction                           \\ \hline

RoBERTa-MIMIC \cite{yang2020clinical}             & EHR  & General RoBERTa    & MIMIC-III Clinical Notes      & Yes             & Clinical Concept Extraction                           \\ \hline

ALBERT-MIMIC \cite{yang2020clinical}               & EHR  & General ALBERTA    & MIMIC-III Clinical Notes      & Yes             & Clinical Concept Extraction                           \\ \hline

DeBERTa-MIMIC \cite{yang2020clinical}              & EHR  & General DeBERTa    & MIMIC-III Clinical Notes      & Yes              & Clinical Concept Extraction                           \\ \hline

Longformer-MIMIC \cite{yang2020clinical}           & EHR  & General Longformer & MIMIC-III Clinical Notes      & Yes               & Clinical Concept Extraction                           \\ \hline

MedBERT \cite{rasmy2021med}                   & EHR  & Scratch            & Private EHR                   & No                & Disease Prediction                                    \\ \hline

BEHRT \cite{li2020behrt}                     & EHR  & Scratch            & Private EHR                   & No               & Disease Prediction                                    \\ \hline

BERT-EHR \cite{meng2021bidirectional}                  & EHR  & Scratch            & Private EHR                   & No                & Disease Prediction                                    \\ \hline

AlphaBERT \cite{chen2020modified}                & EHR  & Scratch            & Private EHR                   & No                & Text Summarization      \\ \hline                               
\end{tabular}}
\end{center}
\caption{\label{table-ehr-based} Summary of EHR-based T-BPLMs.} 
\end{table*}

\section{T-BPLMs TAXONOMY}
\label{taxonomy-sec}

Figure \ref{ammu-taxonomy} shows transformer-based BPLMs taxonomy.

\subsection{ Pretraining Corpus}

\subsubsection{Electronic Health Records}  In the last decade, most hospitals have been using Electronic Health Records (EHRs) to record patient as well as treatment details right from admission to discharge \cite{charles2013adoption}. EHRs contain a vast amount of medical data which can be used to provide better patient care by knowledge discovery and the development of better algorithms. As EHR contains sensitive information related to patients, medical data must be de-identified before sharing. EHRs include both structured and unstructured data \cite{birkhead2015uses,jensen2012mining}. Structured data includes laboratory test results, various medical codes, etc. Unstructured data  include clinical notes like medication instructions, progress notes, discharge summaries, etc. Clinical notes include the most valuable patient information which is difficult and expensive to extract manually. So, there is a need for automatic information extraction methods to utilize the abundant medical data from EHRs in research as well as applications \cite{jensen2012mining,demner2009can,botsis2010secondary}. Following the success of transformer-based PLMs in the general domain, researchers in biomedical NLP also developed EHR-based T-BPLMs by pretraining over clinical notes or medical codes or both. MIMIC \cite{saeed2011multiparameter,johnson2016mimic} is the largest publicly available dataset of medical records. MIT Lab researchers gathered medical records from Beth Israel Deaconess Medical Center, de-identified the sensitive patient information, and then released four versions of the MIMIC dataset.

\begin{table*}[t!]
\begin{center}
{\renewcommand{\arraystretch}{1.5}% for the vertical padding
\begin{tabular}{|p{2.5cm}|p{2cm}|p{1.5cm}|p{3.5cm}|p{1.5cm}|p{4cm}|}
\hline
    \textbf{Model}             & \textbf{Type}      & \textbf{Pretrained from}     & \textbf{Corpus}                    & \textbf{Publicly Available} & \textbf{Evaluation}                       \\ \hline

RadBERT \cite{meng2020self}  & Radiology & General BERT        & RadCore \cite{hassanpour2016information}          & No                 & Radiology Reports Classification \\ \hline

FS-BERT \cite{bressem2020highly} & Radiology & Scratch             & Private Radiology Reports & No                 & Radiology Reports Classification \\ \hline

RAD-BERT \cite{bressem2020highly} & Radiology & German general BERT & Private Radiology Reports & No                 & Radiology Reports Classification \\ \hline
 
\end{tabular}}
\end{center}
\caption{\label{table-radiology-reports} Summary of radiology reports-based T-BPLMs.} 
\end{table*}

\begin{table*}[h]
\begin{center}
{\renewcommand{\arraystretch}{1.5}% for the vertical padding
\begin{tabular}{|p{2.5cm}|p{2cm}|p{1.5cm}|p{3.5cm}|p{1.5cm}|p{4cm}|}
\hline
\textbf{Model} & \textbf{Type}         & \textbf{Pretrained from}   & \textbf{Corpus}                             & \textbf{Publicly Available} & \textbf{Evaluation}                                               \\ \hline

CT-BERT \cite{muller2020covid}          & Social Media & General BERT      & Covid Tweets                       & Yes                & Text Classification                                      \\ \hline

BERTweetCovid19 \cite{nguyen2020bertweet} & Social Media & BERTweet          & Covid Tweets                       & Yes                & Text Classification                                      \\ \hline

BioRedditBERT \cite{basaldella2020cometa}   & Social Media & BioBERT           & Health related Reddit posts        & Yes                & Unsupervised Medical Concept   Normalization             \\ \hline

RuDR-BERT \cite{tutubalina2021russian}        & Social Media & Multilingual BERT & Russian health reviews             & Yes                & Sentence classification and   Clinical Entity Extraction \\ \hline

EnRuDR-BERT \cite{tutubalina2021russian}     & Social Media & Multilingual BERT & Russian and English health reviews & Yes                & ADR (Adverse Drug Reaction) Tweets Classification                                \\ \hline

EnDR-BERT \cite{tutubalina2021russian}    & Social Media & Multilingual BERT & English health reviews             & Yes                & ADR Tweets Classification and ADR   Normalization       \\ \hline
\end{tabular}}
\end{center}
\caption{\label{table-social-media}  Summary of social media-based T-BPLMs.} 
\end{table*}

Alsentzer et al. \cite{alsentzer2019publicly} further pretrained BioBERT (BioBERT-Base v1.0 + PubMed 200K + PMC 270K) on MIMIC III clinical notes to get ClinicalBERT. It took around 17- 18 days to pretrain the model using one GTX TITAN X GPU. It is the first publicly available model pretrained on clinical notes. Si et al. \cite{si2019enhancing} further pre-trained general BERT (base and large) on MIMIC-III clinical notes to get MIMIC-BERT. The authors evaluated the models on various clinical entity extraction datasets. Huang et al. \cite{huang2020clinical} further pre-trained XLNet-base on MIMIC-III Clinical Notes.  The authors released two models namely a) pretrained model on nursing notes and b) pretrained model on discharge summaries. Yang et al. \cite{yang2020clinical} further pre-trained general models like BERT, ELECTRA, RoBERTa, XLNet, and ALBERT on MIMIC-III and released in-domain PLMs. It is the first work to release in-domain models based on all the popular transformer-based PLMs. Unlike the above pre-trained models which are pretrained on clinical text, recent works \cite{yang2020clinical,rasmy2021med,li2020behrt} released models which are pre-trained on disease codes or multi-modal EHR data. BEHRT \cite{li2020behrt} is trained from scratch using 1.6 million patient EHR data with MLM as pretraining task. The authors used code, position, age, and segment embeddings. Med-BERT \cite{rasmy2021med} is trained from scratch using 28,490,650 patient EHR data with MLM and LOS (Length of Stay) as pretraining tasks. The authors used code, serialization and visit embeddings. BERT-EHR \cite{meng2021bidirectional} is trained from scratch using multi-modal data from 43967 patient records with MLM as a pretraining task. Table \ref{table-ehr-based} contains summary of various EHR based BPLMs.

\begin{table*}[t!]
\begin{center}
{\renewcommand{\arraystretch}{1.5}% for the vertical padding
\begin{tabular}{|p{2cm}|p{1.5cm}|p{1.5cm}|p{4cm}|p{1.5cm}|p{4.5cm}|}
\hline
\textbf{Model}    & \textbf{Type}  & \textbf{Pretrained from}   & \textbf{Corpus}   & \textbf{Publicly Available} & \textbf{Evaluation}      \\ \hline

BioBERT \cite{lee2020biobert}           & Scientific Literature & General BERT      & PubMed and PMC  & Yes               & Biomedical NER, RE, and QA.   \\ \hline

RoBERTa-base-PM \cite{lewis2020pretrained}     & Scientific Literature & General RoBERTa   & PubMed and PMC              & Yes     & Sequence Labelling and Text   Classification \\ \hline

RoBERTa-base-PM-Voc \cite{lewis2020pretrained} & Scientific Literature & Scratch           & PubMed and PMC              & Yes    & Sequence Labelling and Text   Classification \\ \hline

BioALBERT \cite{naseem2020bioalbert}           & Scientific Literature & General ALBERT    & PubMed and PMC              & Yes     & Biomedical Concept Extraction     \\ \hline

BioBERTpt-bio \cite{schneider2020biobertpt}      & Scientific Literature & Multilingual BERT & Brazilian Biomedical corpus & No    & Clinical Concept Extraction                  \\ \hline

PubMedBERT \cite{gu2020domain}          & Scientific Literature & Scratch   & PubMed and PMC     & Yes    & BLURB     \\ \hline

BioELECTRA  \cite{ozyurt2020effectiveness}       & Scientific Literature & Scratch           & PubMed   & Yes                & Biomedical NER, QA and RE.         \\ \hline

BioELECTRA ++ \cite{ozyurt2020effectiveness}        & Scientific Literature & BioELECTRA        & PMC                         & Yes     & Biomedical NER, QA and RE.     \\ \hline

BioMegatron \cite{shin2020bio}         & Scientific Literature & Scratch           & PubMed and PMC              & No     & Biomedical NER, RE and QA.    \\ \hline

OuBioBERT \cite{wada2020pre}           & Scientific Literature & Scratch      & PubMed      & Yes    & BLUE    \\ \hline

BlueBERT-PM \cite{peng2019transfer}    & Scientific Literature & General BERT      & PubMed    & Yes    & BLUE     \\ \hline

BioMedBERT \cite{chakraborty2020biomedbert}          & Scientific Literature & General BERT      & BREATHE 1.0                 & No      & Biomedical NER, IR and QA.     \\ \hline

ELECTRAMed \cite{miolo2021electramed}          & Scientific Literature & Scratch           & PubMed                      & Yes     & Biomedical NER, RE and QA                   \\ \hline

BioELECTRA-P \cite{raj2021bioelectra}  & Scientific Literature & Scratch           & PubMed                      & Yes         & BLURB, BLUE       \\ \hline

BioELECTRA-PM \cite{raj2021bioelectra}       & Scientific Literature & Scratch           & PubMed, PMC                 & Yes    & BLURB, BLUE    \\ \hline

BioALBERT-P \cite{naseem2021benchmarking}   & Scientific Literature & ALBERT            & PubMed     & Yes      & BLURB    \\ \hline

BioALBERT-PM \cite{naseem2021benchmarking}  & Scientific Literature & ALBERT            & PubMed, PMC   & Yes   & BLURB   \\ \hline
                                       
\end{tabular}}
\end{center}
\caption{\label{table-scientific-literature}  Summary of scientific literature-based BPLMs. NER - Named Entity Recognition, RE - Relation Extraction, IR - Information Retrieval, QA - Question Answering.} 
\end{table*}

\subsubsection{Radiology Reports} Following the success of EHR-based T-BPLMs, recently researchers focused on developing PLMs specifically for radiology reports. RadCore \cite{hassanpour2016information} dataset consists of around 2 million radiology reports. These reports were gathered from three major healthcare organizations: Mayo Clinic, MD Anderson Cancer Center, and Medical College of Wisconsin in 2007. Meng et al. \cite{meng2020self} further pre-trained general BERT on radiology reports with impression section headings from RadCore dataset to get RadBERT. The authors used RadBERT to classify radiology reports. Bressem et al. \cite{bressem2020highly} released two T-BPLMs for radiology reports namely FS-BERT and RAD-BERT. FS-BERT is obtained by training from scratch using around 3.8 M radiology reports having 415M words (3.6GB) and custom WordPiece vocabulary. RAD-BERT is obtained by further pretraining German general BERT with custom WordPiece vocabulary over 3.8 M radiology reports having 415M words (3.6GB). Table \ref{table-radiology-reports} contains a summary of  radiology reports-based T-BPLMs.

\begin{table*}[t!]
\begin{center}
{\renewcommand{\arraystretch}{1.5}% for the vertical padding
\begin{tabular}{|p{2cm}|p{1.5cm}|p{1.5cm}|p{4cm}|p{1.5cm}|p{4.5cm}|}
\hline
\textbf{Model}   & \textbf{Type}   & \textbf{Pretrained from}       & \textbf{Corpora}   & \textbf{Publicly Available} & {Evaluation}   \\ \hline                                
BlueBERT-P-M3 \cite{peng2019transfer}     & Hybrid & General BERT       & PubMed + MIMIC-III    & Yes   & BLUE    \\ \hline                                     
RoBERTa-base-P-M3 \cite{lewis2020pretrained}    & Hybrid & General RoBERTa     & PubMed + MIMIC-III  & Yes                                                                & Sequence Labelling and Text   Classification \\ \hline

RoBERTa-base-P-M3-Voc \cite{lewis2020pretrained} & Hybrid & Scratch   & PubMed + MIMIC-III   & Yes                                                                & Sequence Labelling and Text   Classification \\ \hline

BioBERTpt-All \cite{schneider2020biobertpt}  & Hybrid & Multilingual BERT     & Brazilian Clinical Text +   Biomedical Text   & Yes                                      & Clinical Concept Extraction                  \\ \hline

BioCharBERT \cite{el2020characterbert}  & Hybrid & General CharacterBERT & PMC Abstracts + MIMIC-III  & Yes                                                                & Clinical NER, MedNLI, RE and STS.        \\ \hline

BERT  (sP +W+BC) \cite{antoun2020arabert}    & Hybrid & Scratch   & PubMed + English Wikipedia +   BooksCorpus    & Yes                                                           & BLUE        \\ \hline

BERT (jpCR+jpW) \cite{antoun2020arabert} & Hybrid & Scratch     & Japanese Clinical Text + Japanese   Wikipedia Text & No                                          & Text Classification        \\ \hline

AraBioBERT \cite{boudjellal2021abioner}  & Hybrid & AraBERT  & General Arabic Text, Arabic Biomedical   Text      & No   & NER                                         \\ \hline

SciBERT \cite{beltagy2019scibert}   & Hybrid & Scratch    & Biomedical + Computer Science   Literature Text    & Yes                & NER and RE                                 \\ \hline

SciFive-P \cite{phan2021scifive}  & Hybrid & T5  & C4, PubMed      & Yes      & NER, RE, MedNLI, QA              \\ \hline

SciFive-PM \cite{phan2021scifive}     & Hybrid & T5      & C4, PubMed, PMC       & Yes       & NER, RE, MedNLI, QA                             \\ \hline

BioALBERT-P-M3 \cite{naseem2021benchmarking} & Hybrid & ALBERT  & PubMed, MIMIC-III    & Yes  & BLURB   \\ \hline                                      
BioALBERT-PM-M3 \cite{naseem2021benchmarking}  & Hybrid & ALBERT   & PubMed, PMC, MIMIC-III   & Yes  & BLURB    \\ \hline                                     
\end{tabular}}
\end{center}
\caption{\label{table-hybrid-corpora}  Summary of hybrid corpora-based T-BPLMs.} 
\end{table*}

\subsubsection{Social Media}  In the last decade, social media has become the first choice for internet users to express their thoughts. Apart from views about general topics, various social media platforms like Twitter, Reddit, AskAPatient, WebMD are used to share health-related experiences in the form of tweets, reviews, questions, and answers \cite{subramanyam2020deep,kalyan2020medical}. Recent works have shown that health-related social media data is useful in many applications to provide better health-related services \cite{o2014pharmacovigilance,limsopatham2015adapting}. The performance of general T-PLMs on Wikipedia and Books Corpus is limited on health-related social media datasets \cite{muller2020covid}. This is because social media text is highly informal with a lot of nonstandard abbreviations, irregular grammar, and typos. Researchers working at the intersection of social media and health, trained social media text-based T-BPLMs to handle social media texts. CT-BERT \cite{muller2020covid} is initialized from BERT-large and further pretrained on a corpus of 22.5M covid related tweets and this model showed up to 30\% improvement compared to BERT-large, on five different classification datasets.  BioRedditBERT \cite{basaldella2020cometa} is initialized from BioBERT and further pretrained on health-related Reddit posts. The authors showed that BioRedditBERT outperforms in-domains models like BioBERT, BlueBERT, PubMedBERT, ClinicalBERT by up to 1.8\% in normalizing health-related entity mentions. RuDR-BERT \cite{tutubalina2021russian} is initialized from Multilingual BERT and pretrained on the raw part of the RuDReC corpus (1.4M reviews). The authors showed that RuDR-BERT outperforms multilingual BERT and Russian BERT on Russian sentence classification and clinical entity extraction datasets by large margins. EnRuDR-BERT \cite{tutubalina2021russian} and EnDR-BERT \cite{tutubalina2021russian} are obtained by further pretraining multilingual BERT on Russian and English health reviews and English health reviews respectively. Table \ref{table-social-media} contains summary of social media text-based BPLMs.

\subsubsection{Scientific Literature} In the last few decades, the amount of biomedical literature is growing at a rapid scale. As knowledge discovery from biomedical literature is useful in many applications, biomedical text mining is gaining popularity in the research community \cite{lee2020biobert}. However, biomedical text significantly differs from the general text with a lot of domain-specific words. As a result, the performance of general T-PLMs is limited in many of the tasks. So, biomedical researchers focused on developing in-domain T-PLMs to handle biomedical text. PubMed and PMC are the two popular sources of biomedical text. PubMed contains only biomedical literature citations and abstracts only while PMC contains full-text biomedical articles. As of March 2020, PubMed includes 30M citations and abstracts while PMC contains 7.5M full-text articles. Due to the large collection and broad coverage, these two are the first choice to pretrain T-BPLMs \cite{lee2020biobert,lewis2020pretrained,naseem2020bioalbert}.

As DSPT is expensive, most of the works developed in-domain T-PLMs by initializing from general BERT models and then further pretraining on biomedical text. BioBERT \cite{lee2020biobert} is the first biomedical pre-trained language model which is obtained by further pretraining general BERT on biomedical literature. BioBERTpt-bio \cite{schneider2020biobertpt}  is obtained by further pretraining BERT Multilingual (base) on Brazilian biomedical corpus - scientific papers from PubMed (0.8M- only literature titles) + Scielo (health – 12.4M + biological- 3.2M: both titles and abstracts). BioMedBERT \cite{chakraborty2020biomedbert} is obtained by further pretraining BERT-large on BREATHE 1.0 corpus. BioMedBERT outperformed BioBERT on biomedical question answering. The key reason for the better performance of BioMedBERT is the diversity of biomedical text in the BREATHE corpus. The main drawback in developing biomedical models by further pretraining general models is the general vocabulary. To overcome this, researchers started to develop biomedical models by using DSPT. Microsoft researchers developed PubMed \cite{gu2020domain} model by DSPT with in-domain vocabulary and whole word masking strategy. OuBioBERT \cite{wada2020pre} is trained from scratch using focused PubMed abstracts (280M words) as core corpora and PubMed abstracts (2800M words) as satellite corpora. It outperforms BioBERT and BlueBERT in many of the tasks in the BLUE benchmark. Table \ref{table-scientific-literature} contains summary of scientific literature-based T-BPLMs.

\subsubsection{Hybrid Corpora}  It is difficult to obtain a large amount of in-domain text in some cases. For example, MIMIC \cite{saeed2011multiparameter,johnson2016mimic} is the largest publicly available dataset of medical records. The MIMIC dataset is small compared to the general Wikipedia corpus or biomedical scientific literature (from PubMed + PMC). However, to pretrain a transformer-based PLM from scratch, we require large volumes of text. To overcome this, some of the models are pretrained on general + in-domain text \cite{boudjellal2021abioner,antoun2020arabert} or, in-domain + related domain text \cite{lewis2020pretrained,peng2019transfer,el2020characterbert,schneider2020biobertpt}. For example, BERT (jpCR+jpW) \cite{antoun2020arabert} - Japanese medical BERT is pretrained from scratch using Japanese Digital Clinical references (jpCR) and Japanese general Wikipedia (jpW). This model outperformed UTH-BERT \cite{kawazoe2020clinical} on Japanese clinical document classification. BERT  (sP +W+BC)\cite{antoun2020arabert} - trained from scratch using small PubMed abstracts as core corpora and English Wikipedia (W) + BooksCorpus(BC) as satellite corpora. This model achieved performance comparable to OuBioBERT \cite{antoun2020arabert} in the BLUE benchmark. AraBioBERT \cite{boudjellal2021abioner} is obtained by further pretraining AraBERT \cite{antoun2020arabert} on general Arabic corpus + biomedical Arabic text. The author showed pretraining a monolingual BERT model like AraBERT on a small-scale domain-specific corpus can still improve the performance of the model. Table \ref{table-hybrid-corpora} contains hybrid corpora-based T-BPLMs.

\subsection{Extensions}
\subsubsection{Language-Specific} Following the success of BioBERT, ClinicalBERT, PubMedBERT in English biomedical tasks, researchers focused on developing T-BPLMs for other languages also by pretraining from scratch \cite{antoun2020arabert,bressem2020highly,akhtyamova2020named} or pretraining from Multilingual BERT \cite{tutubalina2021russian,schneider2020biobertpt} or pretraining from monolingual BERT \cite{bressem2020highly,wang2021cloud,zhang2020conceptualized}  models. For example, CHMBERT \cite{wang2021cloud} is the first Chinese medical BERT model which is initialized from the general Chinese BERT model and further pretrained on a huge (185GB) corpus of Chinese medical text gathered from more than 100 hospitals. MC-BERT \cite{zhang2020conceptualized} is also initialized from general Chinese BERT and further pre-trained on hybrid corpora which includes general, biomedical, and medical texts. Table \ref{table-language-specific} contains a summary of language-specific T-BPLMs.

\begin{table*}[t!]
\begin{center}
{\renewcommand{\arraystretch}{1.5}% for the vertical padding
\begin{tabular}{|p{2cm}|p{1.5cm}|p{1.5cm}|p{4cm}|p{1.5cm}|p{4.5cm}|}
\hline
\textbf{Model}    & \textbf{Language}   & \textbf{Pretrained from}      & \textbf{Corpora}  &  \textbf{Publicly Available} & \textbf{Evaluation}   \\ \hline                                                                                
BERT (jpCR+jpW) \cite{antoun2020arabert} & Japanese   & Scratch  & Japanese Clinical Text + Japanese   Wikipedia           & No                                                                 & Text Classification      \\ \hline                                                                    
BioBERTpt-bio \cite{schneider2020biobertpt}  & Portuguese & Multilingual BERT    & Brazilian Biomedical Text    & Yes                                                                & Clinical Concept Extraction      \\ \hline                                                            
BioBERTpt-clin \cite{schneider2020biobertpt} & Portuguese & Multilingual BERT    & Brazilian Clinical Text                                 & Yes                & Clinical Concept Extraction      \\ \hline                                                            
BioBERTpt-all \cite{schneider2020biobertpt} & Portuguese & Multilingual BERT    & Brazilian Clinical Text +   Biomedical Text             & Yes         & Clinical Concept Extraction     \\ \hline                                                              
RuDR-BERT \cite{tutubalina2021russian}       & Russian    & Multilingual BERT    & Russian Health Reviews    & Yes                           & Text classification and Clinical   NER      \\ \hline

EnRuDR-BERT \cite{tutubalina2021russian}     & Russian    & Multilingual BERT    & Russian and English Health Reviews & Yes                                                                & ADR Tweets Classification     \\ \hline                                                               
FS-BERT \cite{bressem2020highly}         & German     & Scratch              & Private Radiology Reports                               & No  & Radiology Reports Classification           \\ \hline 

RAD-BERT \cite{bressem2020highly}        & German     & General German BERT  & Private Radiology Reports & No                                                                 & Radiology Reports Classification   \\ \hline                                                           
CHMBERT \cite{wang2021cloud}      & Chinese    & General Chinese BERT & Private EHRs   & No                                                                 & Disease Prediction and Department Recommendation     \\ \hline                                        
SpanishBERT \cite{akhtyamova2020named}   & Spanish    & Scratch              & Spanish Biomedical Text   & No                                                                 & Biomedical NER     \\ \hline                                                                 
AraBioBERT \cite{boudjellal2021abioner}     & Arabic     & AraBERT              & General Arabic Text+ Arabic Biomedical Text             & No           & Biomedical NER      \\ \hline   

CamemBioBERT \cite{copara2020contextualized}   & French     & CamemBERT \cite{martin2020camembert}  & French Biomedical Corpus & No                                                                 & Biomedical NER      \\ \hline                                                                          
MC-BERT \cite{zhang2020conceptualized}       & Chinese    & General Chinese BERT & Chinese Biomedical Text, Encyclopedia , Medical records & Yes & ChineseBLUE        \\ \hline      

UTH-BERT \cite{kawazoe2020clinical}       & Japanese   & Scratch  & Japanese Clinical Text                & Yes                                                                & Text Classification     \\ \hline                                                                    
SINA-BERT \cite{taghizadeh2021sina}      & Persian    & ParsBERT \cite{farahani2020parsbert}   & Persian Medical Corpus  & No                      & Medical Question Classification, Medical Question   Retrieval and Medical Sentiment Analysis \\ \hline 

mBERT-Galen \cite{lopez2021transformers}    & Spanish    & Multilingual BERT    & Spanish Clinical Text corpus  & Yes  & Medical Coding                                                                              \\ \hline

BETO-Galen \cite{lopez2021transformers}     & Spanish    & BETO \cite{canete2020spanish}       & Spanish Clinical Text corpus  & Yes              & Medical Coding                                                                    \\ \hline

XLM-R-Galen \cite{lopez2021transformers}    & Spanish    & XLM-R \cite{conneau2020unsupervised}      & Spanish Clinical Text corpus  & Yes                                                                & Medical Coding          \\ \hline                                                                    
\end{tabular}}
\end{center}
\caption{\label{table-language-specific}  Summary of language-specific T-BPLMs.} 
\end{table*}

\subsubsection{Ontology Enriched} T-BPLMs like BioBERT, BlueBERT and PubMedBERT have achieved good results in many biomedical NLP tasks. These models acquire domain-specific knowledge by pretraining on large volumes of biomedical text. Recent works \cite{hao2020enhancing,michalopoulos2021umlsbert,yuan2020coder,liu2021self}  showed that these models acquire only the knowledge available in pretraining corpora and the performance of these models can be further enhanced by integrating knowledge from various human-curated knowledge sources like UMLS. UMLS is a human-curated knowledge source connecting medical terms from various clinical vocabularies. In UMLS, each medical concept has a Concept Unique Identifier (CUI), preferred term, and synonym terms. UMLS concepts are linked by various semantic relationships. Domain knowledge of T-BPLMs can be further enhanced by further pretraining them on UMLS synonyms and relations \cite{hao2020enhancing,michalopoulos2021umlsbert,yuan2020coder,liu2021self} .

\begin{table*}[t!]
\begin{center}
{\renewcommand{\arraystretch}{1.5}% for the vertical padding
\begin{tabular}{|p{2cm}|p{1.5cm}|p{3.5cm}|p{1.5cm}|p{7cm}|}
\hline
\textbf{Model}    & \textbf{Pretrained from}   & \textbf{UMLS data}     & \textbf{Publicly Available} & \textbf{Evaluation}  \\ \hline

Clinical Kb-BERT \cite{hao2020enhancing}   & BioBERT           & UMLS Relations   & Yes & Clinical NER and NLI     \\ \hline                                                                        
Clinical Kb-ALBERT \cite{hao2020enhancing} & General ALBERT    & UMLS Relations   & Yes       & Clinical NER and NLI   \\ \hline                                                                         
UmlsBERT \cite{michalopoulos2021umlsbert}  & ClinicalBERT  & UMLS Synonyms      & Yes  & Clinical NER and NLI      \\ \hline                                                                      
CoderBERT \cite{yuan2020coder}  & BioBERT   & UMLS Synonyms and Relations & Yes  & Unsupervised Medical Concept Normalization, Semantic Similarity, and Relation Classification. \\ \hline

CoderBERT-ALL \cite{yuan2020coder}     & Multilingual BERT & UMLS Synonyms and Relations & Yes & Unsupervised Medical Concept Normalization, Semantic Similarity, and Relation Classification. \\ \hline

SapBERT \cite{liu2021selfsap}  & PubMedBERT        & UMLS Synonyms      & Yes   & Medical Concept Normalization     \\ \hline

SapBERT-XLMR \cite{liu2021selfsap}    & XLM-RoBERTa       & UMLS Synonyms     & Yes  & Medical Concept Normalization  \\ \hline                                                                 

KeBioLM \cite{yuan2021improving}   & PubMedBERT    & UMLS Relations    & Yes   & Named Entity Recognition and Relation   Extraction   \\ \hline                                                                                                           
\end{tabular}}
\end{center}
\caption{\label{table-ontology-enriched}  Summary of ontology enriched T-BPLMs.} 
\end{table*}

Clinical Kb-BERT and Clinical Kb-ALBERT \cite{hao2020enhancing} are obtained by further pretraining BioBERT and ALBERT models on MIMIC-III clinical notes and UMLS relation triplets. Here, pretraining involves three loss functions namely MLM, NSP, and triple classification. Triple classification involves identifying whether two concepts are connected by the relation or not and helps to inject UMLS relationship knowledge into the model. UmlsBERT \cite{michalopoulos2021umlsbert} is initialized from ClinicalBERT and further pretrained on MIMIC-III clinical notes using novel multi-label loss-based MLM and NSP. The novel multi-label loss function allows the model to connect all the words under the same CUI. CoderBERT \cite{yuan2020coder} is initialized from BioBERT and further pre-trained on UMLS concepts and relations using multi-similarity loss and knowledge embedding loss. Multi-similarity loss helps to learn close embeddings for entities under the same CUI and Knowledge embedding loss helps to inject relationship knowledge. SapBERT \cite{liu2021selfsap} is initialized from PubMedBERT and further pre-trained on UMLS synonyms using multi-similarity loss. Table \ref{table-ontology-enriched} contains a summary of ontology enriched T-BPLMs.

\subsubsection{Green Models} CPT allows the general T-PLMs to adapt to in-domain by further pretraining on large volumes of the in-domain corpus. As these models contain vocabulary, which is learned over general text, they cannot represent in-domain words in a meaningful way as many of the in-domain words are split into a number of subwords. This kind of representation increases the overall length of the input as well as hinders the model learning. DSPT or SPT allows the model to have an in-domain vocabulary that is learned over in-domain text. However, both these approaches involve learning the model parameters from scratch which is highly expensive. These approaches being expensive, are far away from the small research labs, and with long runtimes, they are environmentally unfriendly also \cite{poerner2020inexpensive,tai2020exbert}. 

\begin{table*}[t!]
\begin{center}
{\renewcommand{\arraystretch}{1.5}% for the vertical padding
\begin{tabular}{|p{2cm}|p{1.5cm}|p{7cm}|p{3.5cm}|p{1.5cm}|}
\hline
\textbf{Model}   & \textbf{Type}  & \textbf{In-Domain Vocabulary}   & \textbf{Adds}  & {Further pretrained} \\ \hline

GreenBioBERT \cite{poerner2020inexpensive} & Green & Generated using Word2Vec over   biomedical text and then further aligned & In-domain vocabulary                        & No                 \\  \hline

exBERT \cite{tai2020exbert} & Green & Generated using WordPiece over   biomedical text.    & In-domain vocabulary and extension   module & Yes                \\ \hline
                               
\end{tabular}}
\end{center}
\caption{\label{table-green-models}  Summary of Green T-BPLMs.} 
\end{table*}

Recently there is raising interest in the biomedical research community to adapt general T-PLMs models to in-domain in a low cost way and the models contain in-domain vocabulary also \cite{poerner2020inexpensive,tai2020exbert}. These models are referred to as Green Models as they are developed in a low cost environment-friendly approach. GreenBioBERT \cite{poerner2020inexpensive} - extends general BERT to the biomedical domain by refining the embedding layer with domain-specific word vectors. The authors generated in-domain vocabulary using Word2Vec and then aligned them with general WordPiece vectors. With the addition of domain-specific word vectors, the model acquires domain-specific knowledge. The authors showed that GreenBioBERT achieves competitive performance compared to BioBERT in entity extraction. This approach is completely inexpensive as it requires only CPU. exBERT \cite{tai2020exbert} - extends general BERT to the biomedical domain by refining the model with two additions a) in-domain WordPiece vocabulary in addition to existing general domain WordPiece vocabulary b) extension module. The in-domain WordPiece vectors and extension module parameters are learned during pretraining on biomedical text. During pretraining, as the parameters of general BERT are kept fixed, this approach is quite inexpensive. Table \ref{table-green-models} contains summary of Green T-BPLMs.

\subsubsection{Debiased Models} T-PLMs are shown to exhibit bias i.e., the decisions taken by the models may favor a particular group of people compared to others. The main reason for unfair decisions of the models is the bias in the datasets on which the models are trained \cite{meng2021mimic,chen2019can,yu2019framing}. It is necessary to identify and reduce any form of bias that allows the model to take fair decisions without favoring any group. Zhang et al. \cite{zhang2020hurtful} further pretrained SciBERT \cite{beltagy2019scibert} on MIMIC-III clinical notes and showed that the performance of the model is different for different groups. The authors applied adversarial pretraining debiasing to reduce the gender bias in the model. The authors released both the models publicly to encourage further research in debiasing T-BPLMs.

\subsubsection{Multi-Modal Models} T-PLMs have achieved success in many of the NLP tasks including in specific domains like Biomedical. Recent research works have focused on developing pretrained models that can handle multi-modal data i.e., video + text \cite{sun2019videobert,sun2019learning}, image + text  \cite{su2019vl,lu2019vilbert,tan2019lxmert,liu2021medical,monajatipoor2021berthop} etc. In Biomedical domain, models like BERTHop \cite{monajatipoor2021berthop} and Medical-VLBERT \cite{liu2021medical}  have been proposed recently to handle image + text data. BERTHop \cite{monajatipoor2021berthop}  is a multi-modal T-BPLM  developed for Chest X-ray disease diagnosis. Like  ViLBERT \cite{lu2019vilbert} and LXMERT \cite{tan2019lxmert}, BERTHop uses separate encoder to encoder image and text inputs. BERTHop uses PixelHop++ \cite{chen2020pixelhop++} to encode image data and BlueBERT as text encoder. Medical-VLBERT is developed for automatic report generation from COVID-19 scans. Unlike BERTHop, Medical-VLBERT \cite{liu2021medical} uses shared encoder based on VL-BERT \cite{su2019vl} to encode image and text data. 

\section{BIOMEDICAL NLP TASKS}
\label{nlptasks-sec}
\subsection{Natural Language Inference}
Natural Language Inference (NLI) is an important NLP task that requires sentence-level semantics. It involves identifying the relationship between a pair of sentences i.e., whether the second sentence entails or contradicts or neural with the first sentence. Training the model on NLI datasets helps the models to learn sentence-level semantics, which is useful in many tasks like paraphrase mining, information retrieval \cite{reimers2019sentence} in the general domain and medical concept normalization \cite{kalyan2020target,kalyan2020social}, semantic relatedness \cite{kalyan2021hybrid}, question answering \cite{jeong2020transferability} in the biomedical domain. NLI is framed as a three-way sentence pair classification problem. Here models like BERT learn the representation of given two sentences jointly and the three-way task-specific softmax classifier predicts the relationship between the given sentence pair. MedNLI \cite{romanov2018lessons} is the in-domain NLI dataset with around 14k instances generated from MIMIC-III \cite{johnson2016mimic} clinical notes. As MedNLI contains sentence pairs taken from EHRs, Kanakarajan et al. \cite{raj2019saama} further pretrained BioBERT \cite{lee2020biobert} on MIMIC-III and then fine-tuned the model on MedNLI to achieve an accuracy of 83.45\%. Cengiz et al. \cite{cengiz2019ku_ai} applied an ensemble of two BioBERT models and achieved an accuracy of 84.7\%. The authors fine-tuned each of the BioBERT models on general NLI datasets like SNLI \cite{bowman2015large} and MultiNLI \cite{williams2018broad} and then fine-tuned them on MedNLI. 

\subsection{Entity Extraction}
Entity Extraction is the first step in unlocking valuable information in unstructured text data. Entity Extraction is useful in many tasks like entity linking, relation extraction, knowledge graph construction, etc. Extracting clinical entities like drug and adverse drug reactions is useful in pharmacovigilance and biomedical entities like proteins, chemicals and drugs is useful to discover knowledge in scientific literature. Some of the popular entity extraction datasets are I2B2 2010 \cite{uzuner20112010}, VAERS \cite{du2021extracting}, CADEC \cite{karimi2015cadec}, N2C2 2018 \cite{henry20202018} , BC4CHEMD \cite{krallinger2015chemdner}, B5CDR-Chem \cite{li2016biocreative}, JNLPBA \cite{collier2004introduction} and NCBI-Disease \cite{dougan2014ncbi}. 

Most of the existing work approaches entity extraction as sequence labeling or machine reading comprehension. In case of sequence labelling approach, BERT based models generate contextual representations for each token and then softmax layer \cite{johnson2020deidentification,fraser2019extracting,sun2021biomedical}, BiLSTM+Softmax \cite{fraser2019extracting}, BiLSTM+CRF \cite{sun2021biomedical,yu2019biobert,chen2020using,kang2021umls} or CRF \cite{sun2021biomedical,yu2019biobert,portelli2021bert} is applied. Recent works showed adding BiLSTM on the top of the BERT model does not show much difference in performance \cite{yu2019biobert,chen2020using}. This is because transformer encoder layers in BERT based models do the same job of encoding contextual in token representations like BiLSTM. Some of the works experimented with general BERT for extracting clinical and biomedical entities. For example, Portelli et al. \cite{portelli2021bert} showed that SpanBERT+CRF outperformed in-domain BERT models also in extracting clinical entities in social media text. Boudjellal et al. \cite{boudjellal2021abioner}  developed ABioNER by further pretraining AraBERT \cite{antoun2020arabert} on general Arabic corpus + biomedical Arabic text and showed that ABioNER outperformed both multilingual BERT \cite{devlin2019bert} and AraBERT on Arabic biomedical entity extraction. This shows that further pretraining general AraBERT on small in-domain text corpus improves the performance of the model. As in-domain datasets are comparatively small, some of the recent works \cite{sun2021biomedical,miftahutdinov2020biomedical,gao2021pre}  initially fine-tuned the models on similar datasets before fine-tuning on small target datasets. This intermediate fine-tuning allows the model to learn more task-specific knowledge which improves the performance of the model on small target datasets. For example, Gao et al. \cite{gao2021pre} proposed a novel approach entity extraction approach based on intermediate fine-tuning and semi-supervised learning. Here intermediate fine-tuning allows the model to learn more task-specific knowledge while semi-supervised learning allows to leverage task-related unlabelled data by assigning pseudo labels. Recently Sun et al. \cite{sun2021biomedical} formulated biomedical entity extraction as question answering and showed that BioBERT+QA outperformed BioBERT+ (Softmax / CRF / BiLSTM-CRF) on six datasets. 

\subsection{Semantic Textual Similarity}
Semantic Textual Similarity quantifies the degree of semantic similarity between two phrases or sentences. Unlike NLI which classifies the given sentence pair into one of three classes, semantic textual similarity returns a value that indicates the degree of similarity. Both NLI and STS require sentence-level semantics. STS is useful in tasks like concept relatedness \cite{kalyan2021hybrid}, medical concept normalization \cite{kalyan2020target,kalyan2020social} , duplicate text detection \cite{mutinda2020detecting}, question answering \cite{mccreery2019domain,hoogeveen2018detecting} and text summarization \cite{al2016sentence}. Moreover, Reimers et al. \cite{reimers2019sentence} showed that training transformer-based PLMs on STS datasets allow the model to learn sentence-level semantics and hence better represent variable-length texts like phrases or sentences. Models like BERT learn the joint representation of given sentence pair and a task-specific sigmoid layer gives the similarity value. BIOSSES \cite{souganciouglu2017biosses} and Clinical STS \cite{wang20202019} are the commonly used datasets to train and evaluate in-domain STS models. 

Recent works exploited general models for clinical STS \cite{mccreery2019domain,yang2020measurement,wang2020evaluating,wang2020learning}. For example, Yang et al. \cite{yang2020measurement} achieved the best results on the 2019 N2C2 STS dataset using Roberta-large model. As clinical STS datasets are small in size, recent works initially fine-tuned the models on general STS datasets and then fine-tuned them on clinical datasets \cite{mccreery2019domain,yang2020measurement,wang2020learning,mahajan2020identification}. Xiong et al. \cite{xiong2020using} enhanced in-domain BERT-based text similarity using CNN-based character level representations and TransE \cite{bordes2013translating} based entity representations. Mutinda et al. \cite{mutinda2020detecting} achieved a Pearson correlation score of 0.8320 by fine-tuning Clinical BERT on a combined training set having instances from both general and clinical STS datasets. Mahajan et al. \cite{mahajan2020identification} proposed a novel approach based on ClinicalBERT fine-tuned using iterative multi-task learning and achieved the best results on the Clinical STS dataset. Iterative multi-task learning a) helps the model to learn task-specific knowledge from related datasets and b) choose the best-related datasets for intermediate multi-task fine-tuning. The main drawback in the above existing works is giving `[CLS]` vector as sentence pair representation to the sigmoid layer. This is because the `[CLS]` vector contains only partial information. Unlike existing works, Wang et al. \cite{wang2020evaluating} applied hierarchical convolution on the final hidden state vectors and then applied max and min pooling to get the final sentence pair representation and achieved better results. 

\subsection{Relation Extraction}
Relation extraction is a crucial task in information extraction which identifies the semantic relations between entities in text. Entity extraction followed by relation extraction helps to convert unstructured text into structured data. Extracting relations between entities is useful in many tasks like knowledge graph construction, text summarization, and question answering. Some of relation extraction datasets are I2B2 2012 \cite{sun2013evaluating}, AIMED \cite{bunescu2005comparative}, ChemProt \cite{krallinger2017overview}, DDI \cite{herrero2013ddi}, I2B2 2010 \cite{uzuner20112010} and EU-ADR \cite{van2012eu}. Wei et al. \cite{wei2019relation} achieved the best results on two datasets using MIMIC-BERT \cite{si2019enhancing} +Softmax. Thillaisundaram and Togia \cite{thillaisundaram2019biomedical} applied SciBERT +Softmax to extract relations from biomedical abstracts as part of AGAC track of BioNLP-OST 2019 shared tasks \cite{wang2019overview}. Liu et al. \cite{liu2020document} proposed SciBERT+Softmax for relation extraction in biomedical text. They showed SciBERT+Softmax outperforms BERT+Softmax on three biomedical relation extraction datasets. The main drawback in the above existing works is that they utilize only the partial knowledge from the last layer in the form of `[CLS]` vector. Su et al. \cite{su2020investigation} added attention on the top of BioBERT to fully utilize the information in the last layer and achieved the best results on three biomedical extraction datasets. The authors generated the final representation by concatenating `[CLS]` vector and weighted sum vector of final hidden state vectors. When compared to applying LSTM on the final hidden state vectors, the attention layer gives better results. 

\subsection{Text Classification}
Text classification involves assigning one of the predefined labels to variable-length texts like phrases, sentences, paragraphs, or documents. Text classification involves an encoder which is usually a transformer-based PLM and a task-specific softmax classifier. `[CLS]` vector or weighted sum of final hidden state vectors is treated as an aggregate representation of given text. The fully connected dense layer in the classifier projects text representation vector into n-dimensional vector space where ‘n’ represents the number of predefined labels. Finally, the softmax function is applied to get the probabilities of all the labels. Garadi et al. \cite{al2021text} formulated prescription medication (PM) identification from tweets as four-way text classification. They achieved good results using models like BERT, RoBERTa, XLNet, ALBERT, and DistillBERT. They trained different ML-based meta classifiers with predictions from pre-trained models as inputs and further improved the results.

Shen et al. \cite{shen2021extracting} applied various in-domain BERT for Alzheimer disease clinical notes classification and achieved the best results using PubMedBERT and BioBERT. They generated labels for the training instances using a rule-based NLP algorithm.  Chen et al. \cite{chen2019hitsz} further pretrained the general BERT model on 1.5 drug-related tweets and showed that further pretraining improves the performance of the model on ADR tweets classification. Recent works \cite{tang2019progress,wood2020automated} showed that adding attention on the top of the BERT model improves the performance of the model in clinical text classification. They introduced a custom attention model to aggregate the encoder output and showed that this leads to improved performance and interpretability.

\begin{table*}[t!]
\begin{center}
{\renewcommand{\arraystretch}{1.5}% for the vertical padding
\begin{tabular}{|p{3.5cm}|p{1.5cm}|p{1.5cm}|p{1.5cm}|p{1.5cm}|p{1.5cm}|p{1.5cm}|p{1.5cm}|}
\hline
\textbf{Model}   & \textbf{NER}  & \textbf{PICO}   & \textbf{RE} & \textbf{SS} & \textbf{DC} & \textbf{QA} & \textbf{BLURB Score} \\ \hline
BioELECTRA \cite{raj2021bioelectra} & 86.67	& 74.13	& 81.44	& 92.76	& 84.20 & 	76.38 & 82.60 \\ \hline
PubMedBERT \cite{gu2020domain} & 86.13	& 73.72	& 80.59 &	92.31 &	82.62	& 73.61 & 81.50 \\ \hline
BioBERT \cite{lee2020biobert} & 85.81	& 73.18 & 	79.79 &	89.52 &	81.54 &	72.19 & 80.34 \\ \hline
Scibert \cite{beltagy2019scibert} & 85.43 &	73.12 &	79.56 &	86.25	& 80.66 &	68.12 &  78.86 \\ \hline
ClinicalBERT \cite{alsentzer2019publicly} & 83.99 &	72.06 &	76.91 &	91.23 &	80.74 &	58.79 & 77.29 \\ \hline
BlueBERT \cite{peng2019transfer} & 84.50 &	72.54 &	76.13 &	85.38 &	80.48 &	58.57 & 76.27 \\ \hline

\end{tabular}}
\end{center}
\caption{\label{table-evaluation}  Performance comparison of various T-BPLMs (scores from BLURB benchmark). NER- Named Entity Recognition, PICO - Patient Population Interventions Comparator and Outcomes, RE-Relation Extraction, SS - Sentence Similarity, DC - Document Classification and QA - Question Answering. } 
\end{table*}

\subsection{Question Answering}
Question Answering (QA) aims to extract answers for the given queries. QA helps to quickly find information in clinical notes or biomedical literature and thus saves a lot of time. The main challenge in developing automated QA systems for the clinical or biomedical domain is the small size of datasets. Developing large QA datasets in the clinical or biomedical domain is quite expensive and requires a lot of time also. Some of the popular in-domain QA datasets are emrQA \cite{pampari2018emrqa}, CliCR \cite{suster2018clicr}, PubMedQA \cite{jin2019pubmedqa} COVID-QA \cite{moller2020covid}, MASH-QA \cite{zhu2020question} and Health-QA \cite{zhu2019hierarchical}. Chakraborty et al. \cite{soni2020evaluation} showed BioMedBERT obtained by further pretraining BERT-Large on BREATHE 1.0 corpus outperformed BioBERT on biomedical question answering. The main reason for this is the diversity of text in BREATHE 1.0 corpus. 

Pergola et al. \cite{pergola2021boosting} introduced Biomedical Entity Masking (BEM) which allows the model to learn entity-centric knowledge during further pretraining. They showed that BEM improved the performance of both general and in-domain models on two in-domain QA datasets. Recent works used intermediate fine-tuning on general QA \cite{soni2020evaluation,yoon2019pre} or NLI \cite{jeong2020transferability} datasets or multi-tasking \cite{akdemir2020transfer} to improve the performance of in-domain QA models. For example, Soni et al. \cite{soni2020evaluation} achieved the best results on a) CliCR by intermediate fine-tuning on SQuaD using BioBERT and b) emrQA by intermediate fine-tuning on SQuaD and CliCR using Clinical BERT. Yoon et al. \cite{yoon2019pre} showed that intermediate fine-tuning on general domain Squad datasets improves the performance on biomedical question answering datasets. Akdemir et al. \cite{akdemir2020transfer} proposed a novel  multi-task model based on BioBERT for biomedical question answering. They used biomedical NER as an auxiliary task and showed that transfer learning from the bioNER task improves performance on question answering tasks.

\subsection{Text Summarization}
In general, sources of biomedical information like clinical notes, scientific papers, radiology reports are length in nature. Researchers and domain experts need to go through a number of biomedical documents. As biomedical documents are length in nature, there is a need for automatic biomedical text summarization which reduces the effort and time for researchers and domain experts \cite{mishra2014text,moradi2018different}. Text summarization falls into two broad categories namely extractive summarization which identifies the most relevant sentences in the document while abstractive summarization generates new text which represents the summary of the document \cite{gigioli2018domain}. There are no standard datasets for biomedical text summarization. Researchers usually treat scientific papers as documents and their abstracts as summaries \cite{moradi2020deep,moradi2020summarization}.

Moradi et al. \cite{moradi2020deep} proposed a novel approach to summarize biomedical scientific articles. They embedded sentences, generated clusters, and then extracted the most informative sentences from each of the clusters. They showed that BERT-large outperformed other models including the in-domain BERT models like BioBERT. In the case of small models, BioBERT outperformed others. Moradi et al. \cite{moradi2020summarization} proposed a novel approach based on word embeddings and graph ranking to summarize the biomedical text. They generated a graph with sentences as nodes and edges as relations whose strength is measured by cosine similarity between sentence vectors generated by averaging BioBERT and Glove embeddings and finally, graph ranking algorithms identify the important sentences. Du et al. \cite{du2020biomedical} introduced a novel approach called BioBERTSum to summarize the biomedical text. BioBERTSum uses BioBERT to encode sentences, transformer decoder + sigmoid to calculate the scores for each sentence. The sentences with the highest score are considered as the summary. Chen et al. \cite{chen2020modified} proposed a novel clinical text summarization based on AlphaBERT. 

\section{Evaluation}
\label{evaluation-sec}
Benchmarks are useful to evaluate the progress in pretrained models. GLUE is the first benchmark proposed to evaluate pretrained models. Following GLUE, a number of benchmarks are proposed in general NLP. Inspired by the benchmarks in general NLP, Biomedical  research community proposed benchmarks like BLUE, BLURB and CBLUE. We summarize the performance of various T-BPLMs in Table \ref{table-evaluation}.

\section{CHALLENGES AND SOLUTIONS}
\label{challenges-sec}
\begin{table*}[t!]
\begin{center}
{\renewcommand{\arraystretch}{1.5}% for the vertical padding
\begin{tabular}{|p{4cm}|p{4cm}|p{4cm}|p{4cm}|}
\hline
Approach     & Description   & Pros       & Cons   \\ \hline

Intermediate Fine-Tuning & Model is fine-tuned on source   dataset before fine-tuning on target dataset.    & Allows the model to gain domain or   task-specific knowledge.     & Requirement of labeled datasets.   \\ \hline

Multi-Task Fine-tuning   & Model is fine-tuned on multiple   tasks simultaneously.                          & Allow the model to learn from   multiple tasks simultaneously.    & Requirement of labeled   datasets. Fine-tuning must be done iteratively to identify the best subset of   tasks.   \\ \hline

Data Augmentation        & Augment the training set using Back   Translation or EDA techniques.             & Very simple and easy to implement. & Label preserving is not guaranteed.    \\ \hline

Semi-Supervised Learning & Fine-tunes the model on training   instances along with pseudo labeled instances & Allows the model to leverage task-related   unlabelled instances. & Fine-tuning must be done   iteratively to reduce the noisy labeled instances.       \\ \hline                            
\end{tabular}}
\end{center}
\caption{\label{table-small-datasets}  Summary of various approaches to handle small biomedical datasets using t-BPLMs.} 
\end{table*}

\subsection{Low Cost Domain Adaptation}
The two popular approaches for developing T-BPLMs are MDPT and DSPT. These approaches involve pretraining on large volumes of in-domain text using high-end GPUs or TPUs for days. These two approaches are quite successful in developing BPLMs. However, these approaches are quite expensive requiring high computing resources with long pretraining durations \cite{poerner2020inexpensive}. For example, BioBERT - it took around ten days to adapt general BERT to the biomedical domain using eight GPUs \cite{lee2020biobert}. Moreover, DSPT is more expensive compared to continual pretraining as it involves learning model weights from scratch \cite{poerner2020inexpensive,tai2020exbert}.  So, there is a need for lost cost domain adaptation methods to adapt general BERT models to the biomedical domain. Two such low-cost domain adaptation methods are \textbf{a)} \textbf{Task Adaptative Pretraining (TAPT)} - It involves further pretraining on task-related unlabelled instances \cite{gururangan2020don}. TAPT allows the models to learn domain as well as task-specific knowledge by pretraining on a relatively small number of task-related unlabelled instances. The number of unlabelled instances can be increased by retrieving task-related unlabelled sentences using lightweight approaches like VAMPIRE \cite{gururangan2019variational}.  \textbf{b)} \textit{Extending embedding layer} - General T-PLMs can be adapted to the biomedical domain by refining the embedding layer with the addition of new in-domain vocabulary \cite{poerner2020inexpensive,tai2020exbert}. The new in-domain vocabulary can be i) generated over in-domain text using Word2Vec and then aligned with general word-piece vocabulary \cite{poerner2020inexpensive} or ii) generated over in-domain text using word-piece \cite{tai2020exbert}. 

\subsection{ Ontology Knowledge Injection}
Models like BioBERT, PubMedBERT have achieved good results in many of the tasks. However, these models lack knowledge from human-curated knowledge sources. These models can be further enhanced by ontology knowledge injection. Ontology knowledge injection can be done in many ways namely a) continual pretraining the models on UMLS synonyms \cite{michalopoulos2021umlsbert,liu2021self} or relations \cite{hao2020enhancing} or both \cite{yuan2020coder} b) continual pretraining the models on UMLS concept definitions \cite{mao2020use} and c) feature vector constructed using ontology is added to the  sequence vector learned by the models \cite{chen2019hitsz}.

\subsection{ Small Datasets}
Pretraining on large volumes of in-domain text allows the model language representations while fine-tuning allows the model to learn task-specific knowledge. During fine-tuning the model must learn sufficient task-specific knowledge to achieve good results on the task where the input distribution, as well as label space, is different from pretraining \cite{phang2018sentence,pruksachatkun2020intermediate}. With small target datasets, the models are not able to learn enough task-specific which limits the performance. To over the small size of target datasets, the possible solutions are

\textbf{Intermediate Fine-Tuning} – Intermediate fine-tuning on large, related datasets allows the model to learn more domain or task-specific knowledge which improves the performance on small target datasets \cite{cengiz2019ku_ai,sun2021biomedical,gao2021pre,mccreery2019domain,yang2020measurement,wang2020learning,yoon2019pre,jeong2020transferability} and 

\textbf{Multi-Task Fine-Tuning} -  Multi-task fine-tuning allows the model to be fine-tuned on multiple tasks simultaneously \cite{liu2019multi,zhang2021survey,khan2020mt}. Here the embedding and transformer encoder layers are common for all the tasks and each task has a separate task-specific layer. Multi-task fine-tuning allows the model to gain domain as well as task-specific reasoning knowledge from multiple tasks. Multi-Task fine-tuning is more useful in low resource scenarios which are common in the biomedical domain \cite{mahajan2020identification,peng2020empirical,khan2020mt,mulyar2020mt}.

\textbf{Data Augmentation} - Data augmentation helps us to create new training instances from existing instances. These newly creating training instances are close to original training data and helpful in low resource scenarios. Back translation and EDA \cite{wei2019eda} are the top popular techniques for data augmentation. For example, Wang et al. \cite{wang2020evaluating} used back translation to augment the training instances to train the clinical text similarity model. The domain-specific ontologies like UMLS can also be used to augment the training instances \cite{kang2021umls}. 

\textbf{Semi-Supervised Learning} - Semi-supervised learning augments the training set with pseudo-labeled instances. The model which is fine-tuned on the original training set is used to label the task-related unlabelled instances. The model is again fine-tuned on the augmented training set and this process is repeated until the model converges \cite{gao2021pre,yu2020robust} . Table \ref{table-small-datasets} contains a brief summary of these approaches.

\subsection{Robustness to Noise}
Transformed based PLMs have achieved the best results in many of the tasks. However, the performance of these models on noisy test instances is limited  \cite{jin2020bert,pruthi2019combating,kalyan2021bertmcn,araujo2020adversarial}. This is because the model is mostly trained on less noisy instances. As these models mostly never encounter noisy instances during fine-tuning, the performance is significantly reduced on noisy instances. Apart from this, the noisy words in the instances are split into several subtokens which affect the model learning. The robustness of models is crucial particularly insensitive domains like biomedical \cite{kalyan2021bertmcn,araujo2020adversarial}. Two possible solutions are \textbf{a)} CharBERT \cite{el2020characterbert} – replaced the WordPiece based embedding layer with CharCNN based embedding layer. Here word representation is generated from character embeddings using CharCNN. \textbf{b)} Adversarial Training \cite{araujo2020adversarial} – Here, the training set is augmented with the noisy instances. Training the model on an augmented training set exposes the model to noisy instances and hence the model performs better on noisy instances. 

\subsection{Quality In-Domain Word Representations}
Continual pretraining allows the general T-PLMs to adapt to in-domain by further pretraining on large volumes of in-domain text. Though the models are adapted to in-domain, they still contain general vocabulary. As the vocabulary is learned over general text, it mostly includes subwords and words which are specific to the general domain. As a result, many of the in-domain words are not represented in a meaningful way. The two possible options to represent in-domain words in a meaningful way are \textbf{a)} in-domain vocabulary through DSPT \cite{gu2020domain} \textbf{b)} extending the general vocabulary with in-domain vocabulary \cite{poerner2020inexpensive,tai2020exbert}. 

\subsection{Low Resource (In-Domain Corpus) Pretraining}
CPT or DSPT involves pretraining the language model on large volumes of in-domain text. During pretraining, the model learns language representations that are useful across many tasks. The size of the pretraining corpus influences how well the model learns the language representations. It is not possible to get a large volume of in-domain text all the time. In such scenarios with less in-domain corpus, the model may not learn well when trained using any of the above two methods. The possible solution for this is simultaneous pretraining. In simultaneous pretraining \cite{wada2020pre}, the model is trained on combined corpora having both general and in-domain text. As the in-domain text is comparatively less, up-sampling can be used to have a balanced pretraining.

\subsection{Quality Sequence Representation}
For text classification or sentence pair classification tasks like NLI and STS, Devlin et al. \cite{devlin2019bert} suggested to use the final hidden vector of the special token added at the beginning of the input sequence as the final input sequence representation. According to Devlin et al. \cite{devlin2019bert}, the final hidden vector of the special token aggregates the entire sequence information. The final hidden vector is given to a linear layer which projects into n-dimensional vector space whether n represents the size of label space. Finally, a softmax is applied to convert it into a vector of probabilities. However, some of the recent works showed that involving all the final hidden vectors using max-pooling \cite{reimers2019sentence}, attention \cite{su2020investigation,tang2019progress,wood2020automated}, or hierarchical convolution layers \cite{wang2020evaluating,wang2020learning} gives a much better final sequence representation compared to using only special token vector.  

\section{FUTURE DIRECTIONS}
\label{future-sec}
\subsection{Mitigating Bias}
With the success of deep learning models in various tasks, deep learning-based systems are used to automate the decisions like department recommendation \cite{wang2021cloud}, disease prediction \cite{meng2021bidirectional,rasmy2021med} etc. However, these models are shown to exhibit bias i.e., the decisions taken by the models may favor a particular group of people compared to others. The main reason for unfair decisions of the models is the bias in the datasets on which the models are trained \cite{meng2021mimic,chen2019can,yu2019framing}. Real-world datasets have a bias in many forms. It can be based on various attributes like gender, age, ethnicity, and marital status. These attributes are considered as protected or sensitive \cite{mehrabi2021survey}.  For example, in the MIMIC-III dataset \cite{johnson2016mimic} a) heart disease is more common in males compared to females– an example of gender bias b) there are fewer clinical studies involving black patients compared to other groups –an example of ethnicity bias.  It is necessary to identify and reduce any form of bias that allows the model to take fair decisions without favoring any group. There are few works that identified and addressed bias in transformer-based biomedical language models. Zhang et al. \cite{zhang2020hurtful} using a simple word competition task showed that SciBERT \cite{beltagy2019scibert} exhibits ethnicity bias. Moreover, the authors showed SciBERT model when further-pre-trained on clinical notes exhibits performance differences for different protected attributes. They further showed that adversarial pretraining debiasing has little impact in reducing bias. Minot et al. \cite{minot2021interpretable} proposed an approach based on data augmentation to identify and reduce gender bias in patient notes. This is an area that needs to be explored further to improve reduce bias and improve the fairness in model decisions.

\subsection{Privacy Issues}
Every patient visit is recorded in the clinical records. Apart from patient visits, clinical records contain the past and the present medical history of the patient. Such sensitive data should not be disclosed as it may harm the patients physically or mentally \cite{nakamura2020kart}. Usually, the clinical records are shared for research purposes only after de-identifying the sensitive information. However, it is possible to recover sensitive patient information from the de-identified medical records. Recent works showed that there is data leakage from pre-trained models in the general domain i.e., it is possible to recover personal information present in the pretraining corpora \cite{misra2019black,hisamoto2020membership}. Due to data leakage, the models pre-trained on proprietary corpora, cannot be released publicly. Recently, Nakamura et al. \cite{nakamura2020kart} proposed KART framework which can conduct various attacks to assess the leakage of sensitive information from pre-trained biomedical language models. We strongly believe there is a need for more work in this area to assess as well as address the data leakage in biomedical language models. 

\subsection{Domain Adaptation}
In the beginning, the standard approach to develop BPLMs is to start with general PLMs and then further pretrain them on large volumes of biomedical text \cite{lee2020biobert}. The main drawback of this approach is the lack of in-domain vocabulary. Without domain-specific vocabulary, many of the in-domain are split into a number of subwords which hinders model learning during pretraining or fine-tuning. Moreover, continual pretraining is quite expensive as it involves pretraining on large volumes of unlabeled text. To overcome these drawbacks, there are low-cost domain adaptation approaches that extend the general domain vocabulary with in-domain vocabulary \cite{poerner2020inexpensive,tai2020exbert}. The extra in-domain vocabulary is generated using Word2vec and then aligned \cite{poerner2020inexpensive} or generated directly using WordPiece \cite{tai2020exbert} over biomedical text.  The main drawback in these low-cost domain adaptation approaches is an increase in the size of the model with the addition of in-domain vocabulary. Further research on this topic can result in more novel methods for low-cost domain adaptation.

\subsection{Novel Pretraining Tasks}
Most of the biomedical language models (except ELECTRA-based models) are pre-trained using MLM. In MLM, only 15\% of tokens are randomly masked and the model learns by predicting that 15\% of masked tokens only. Here the main drawbacks are a) as tokens are randomly chosen for masking, the model may not learn much by predicting random tokens b) as only 15\% of tokens are predicted, the training signal per example is less. So, the model has to see more examples to learn enough language information which results in the requirement of large pretraining corpora and more computational resources. There is a need for novel pretraining tasks like Replaced Token Detection (RTD) which can provide more training signal per example. Moreover, when the model is pretrained using multiple pretraining tasks, the model receives more training signals per example and hence can learn enough language information using less pretraining corpora and computational resources \cite{aroca2020losses}. 

\subsection{Benchmarks}
In general, a benchmark is a tool to evaluate the performance of models across different NLP tasks. A benchmark is required because we expect the pre-trained language models to be general and robust i.e., the models perform well across tasks rather than on one or two specific tasks. A benchmark with one or more datasets for multiple NLP tasks helps to assess the general ability and robustness of models. In general domain, we have a number of benchmarks like GLUE \cite{wang2018glue} and SuperGLUE \cite{wang2019superglue} (general language understanding), XGLUE \cite{liang2020xglue} (cross lingual language understanding) and LinCE \cite{aguilar2020lince} (code switching).  In biomedical domain there are three benchmarks namely BLUE \cite{peng2019transfer}, BLURB \cite{gu2020domain} and ChineseBLUE \cite{zhang2020conceptualized}. BLUE introduced by Peng et al. \cite{peng2019transfer} contains ten datasets for five biomedical NLP tasks, while BLURB contains thirteen datasets for six tasks and ChineseBLUE contains eight tasks with nine datasets. BLUE and ChineseBLUE include both EHR and scientific literature-based datasets, while BLURB includes only biomedical scientific literature-based datasets. The semantics of EHR and medical social media texts are different from biomedical scientific literature. So, there is a need of exclusive benchmarks for EHR and medical social media-based datasets.  

\subsection{Intrinsic Probes}
During pretraining, PLMs learn syntactic, semantic knowledge along with factual and common-sense knowledge available in the pretraining corpus \cite{kalyan2021ammus}. Intrinsic probes through light on the knowledge learned by PLMs during pretraining. In general NLP, researchers proposed several intrinsic probes like LAMA, Negated and Misprimed LAMA \cite{petroni2019language}, XLAMA \cite{kassner2021multilingual}, X-FACTR \cite{jiang2020x}, MickeyProbe \cite{lin2021common} to understand the knowledge encoded in pretrained models. For example, LAMA \cite{petroni2019language} probes the factual and common-sense knowledge of English pretrained models, while X-FACTR \cite{jiang2020x} probes the factual knowledge of multi-lingual pretrained models. However, there is no such intrinsic probes in Biomedical domain to through light on the knowledge learned by BPLMs during pretraining. This is an area which requires much attention from Biomedical NLP community.

\subsection{Efficient Models}
Pretraining provides BPLMs with necessary background knowledge which can be transferred to downstream tasks. However, pretraining is computationally very expensive and also requires large volumes of pretraining data. So, there is need of novel model architecture which reduces the pretraining time as well as the amount of pretraining corpus. In general NLP, recently efficient models like ConvBERT \cite{jiang2020convbert} and DeBERTa \cite{he2020deberta} are proposed which reduces the pretraining time and amount of pretraining corpus required respectively. DeBERTa with two novel improvements (Disentangled attention mechanism and enhanced masked decoder) achieves better compared to RoBERTa. DeBERTa is pretrained on just 78GB of data while RoBERTa is pretrained on 160GB of data. ConvBERT with mixed attention block outperforms ELECTRA while using just $1/4^{th}$ of its pretraining cost. Biomedical NLP research community must focus on developing pretrained models based on these novel model architectures. 

\section{Limitations}
We have comprehensively covered the research works related to T-BPLMs. As our focus is on T-BPLMs, we have not included any papers related to context insensitive biomedical embeddings. For detailed information regarding context insensitive biomedical embeddings, please refer the survey paper written by Kalyan and Sangeetha \cite{kalyan2020secnlp}.  As it is a survey focused on T-BPLMs, we have covered the foundation concepts like transformers and self-supervised learning in a very brief way only. 

\section{Conclusion}
Here, we present the recent trends in transformer-based BPLMs. We explain various core concepts like pretraining methods, pretraining tasks, fine-tuning methods and embedding types. We present a taxonomy for transformer-based BPLMs. Finally, we discuss some of the challenges and possible solutions and finally conclude with a discussion on open issues.

\ifCLASSOPTIONcompsoc
  \section*{Acknowledgments}
\else
  \section*{Acknowledgment}
\fi

Kalyan would like to thank his father Katikapalli Subramanyam for giving a) \$750 to buy a new laptop, 24-inch monitor and study table.  b) \$180 for one year subscription of Medium, Overleaf and Edraw MindMaster software. Edraw MindMaster is used to create all the diagrams in the paper.

\ifCLASSOPTIONcaptionsoff
  \newpage
\fi

\bibliographystyle{IEEEtran}
\bibliography{ammu-arxiv.bib}

\end{document}